\documentclass[10pt,twocolumn,letterpaper]{article}

\usepackage{cvpr}
\usepackage{times}
\usepackage{epsfig}
\usepackage{graphicx}
\usepackage{amsmath}
\usepackage{amssymb}

\usepackage{subfigure}
\usepackage{booktabs}
\usepackage{multirow}
\newcommand{\ms}[1]{\tiny{$\pm$#1}}
\usepackage{algorithm}
\usepackage{algpseudocode}

\usepackage{diagbox}
\usepackage{multicol}
\usepackage[pagebackref=true,breaklinks=true,letterpaper=true,colorlinks,bookmarks=false]{hyperref}

\cvprfinalcopy % *** Uncomment this line for the final submission

\def\cvprPaperID{8349} % *** Enter the CVPR Paper ID here

\begin{document}

%%%%%%%%% TITLE
\title{Regularizing Class-wise Predictions via Self-knowledge Distillation}

% \author{Sukmin Yun\\
% KAIST\\
% Daejeon, Korea\\
% {\tt\small sukmin.yun@kaist.ac.kr}
% % \hspace{-0.1in}
% \and
% Jongjin Park\\
% KAIST\\
% Daejeon, Korea\\
% {\tt\small jongjin.park@kaist.ac.kr}
% \hspace{-0.1in}
% \and
% Kimin Lee\footnote{work done while at KAIST}\\
% UC Berkeley\\
% Berkeley, CA, USA\\
% {\tt\small kiminlee@berkeley.edu}
% % \hspace{-0.1in}
% \and
% Jinwoo Shin\\
% KAIST\\
% Daejeon, Korea\\
% {\tt\small jinwoos@kaist.ac.kr}
% }

\author{Sukmin Yun$^{1}\thanks{Equal contribution}$\quad Jongjin Park$^{1}$\footnotemark[1]\quad Kimin Lee$^{2}$\thanks{Work was done while the author was at KAIST}\quad Jinwoo Shin$^{1}$\\
$^{1}$Korea Advanced Institute of Science and Technology, South Korea\\
$^{2}$University of California, Berkeley, USA\\
{\tt\small \{sukmin.yun, jongjin.park, jinwoos\}@kaist.ac.kr} \quad
{\tt\small kiminlee@berkeley.edu}\\
}

\maketitle
%\thispagestyle{empty}

%%%%%%%%% ABSTRACT
\begin{abstract}
Deep neural networks with millions of parameters may suffer from poor {generalization} due to overfitting.
To mitigate the issue,
we propose a new regularization method that penalizes the predictive distribution between similar samples.
In particular,
we distill the predictive distribution between different samples of the same label %and %augmented samples of the same source 
during training.
This results in regularizing the dark knowledge (i.e., the knowledge on wrong predictions) %of DNN
of a single network (i.e., a self-knowledge distillation) by forcing it to produce more meaningful and consistent predictions in a class-wise manner.
% In this way, our approach achieves two goals that have been developed in independent literature: preventing overconfident predictions and reducing the intra-class variations.
Consequently, it %Specifically, it has the advantage of preventing 
mitigates overconfident predictions and reduces intra-class variations.
Our experimental results on various image classification tasks
demonstrate that %effectiveness of 
the simple yet powerful method can 
significantly improve not only the generalization ability but also the calibration performance of modern convolutional neural networks. 
\end{abstract}

%%%%%%%%% BODY TEXT
\vspace{-0.05in}
\section{Introduction}
%\vspace{-0.05in}
Deep neural networks (DNNs) have achieved state-of-the-art performance on many computer vision tasks, 
\eg, image classification \cite{he2016deep}, generation \cite{brock2018large}, and segmentation \cite{he2017mask}.
As the scale of training dataset increases, 
the size of DNNs (\textit{i.e.}, the number of parameters) also scales up to handle such a large dataset efficiently.
However, networks with millions of parameters may incur overfitting and suffer from poor generalizations \cite{pereyra2017regularizing, zhang2016understanding}.
To address the issue, % of DNNs,
many regularization strategies have been investigated in the literature:
{early stopping} \cite{bishop1995regularization}, $L_1$/$L_2$-regularization \cite{nowlan1992simplifying}, dropout \cite{srivastava2014dropout}, batch normalization \cite{ioffe2015batch} and data augmentation \cite{cubuk2018autoaugment}.

\begin{figure}[t]
\vspace{-0.15in}
\centering
\subfigure[{Overview of} our regularization scheme]
{
\includegraphics[width=0.45\textwidth]{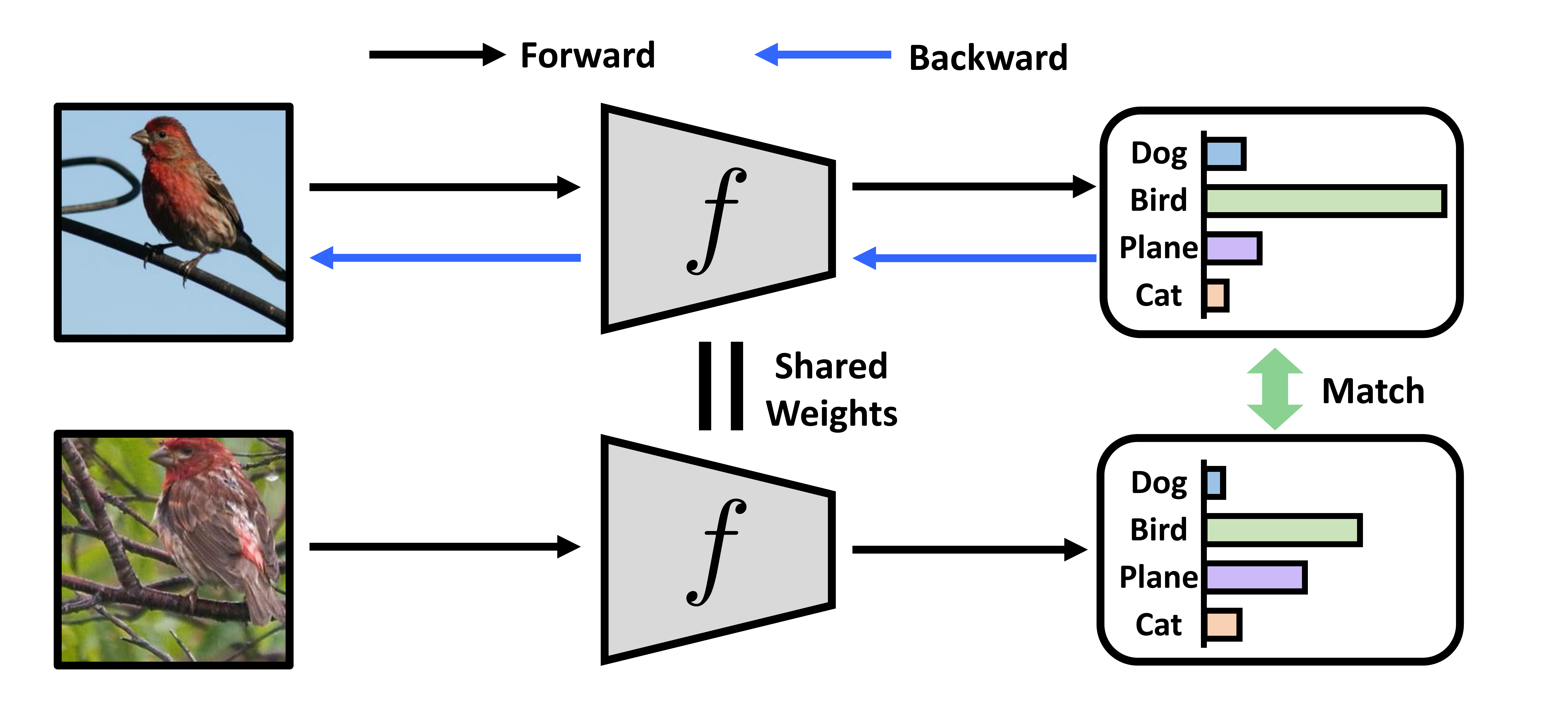}\label{fig:cskd}} 
\vspace{-0.15in}
\,
% \subfigure[Cross-entropy]
% {
% \includegraphics[width=0.2\textwidth]{figures/tSNE_c100_ce.pdf}\label{fig:tsne_ce}} 
% \,
\subfigure[\textcolor{black}{Top-5 softmax scores on misclassified samples}] 
{
\includegraphics[width=0.45\textwidth]{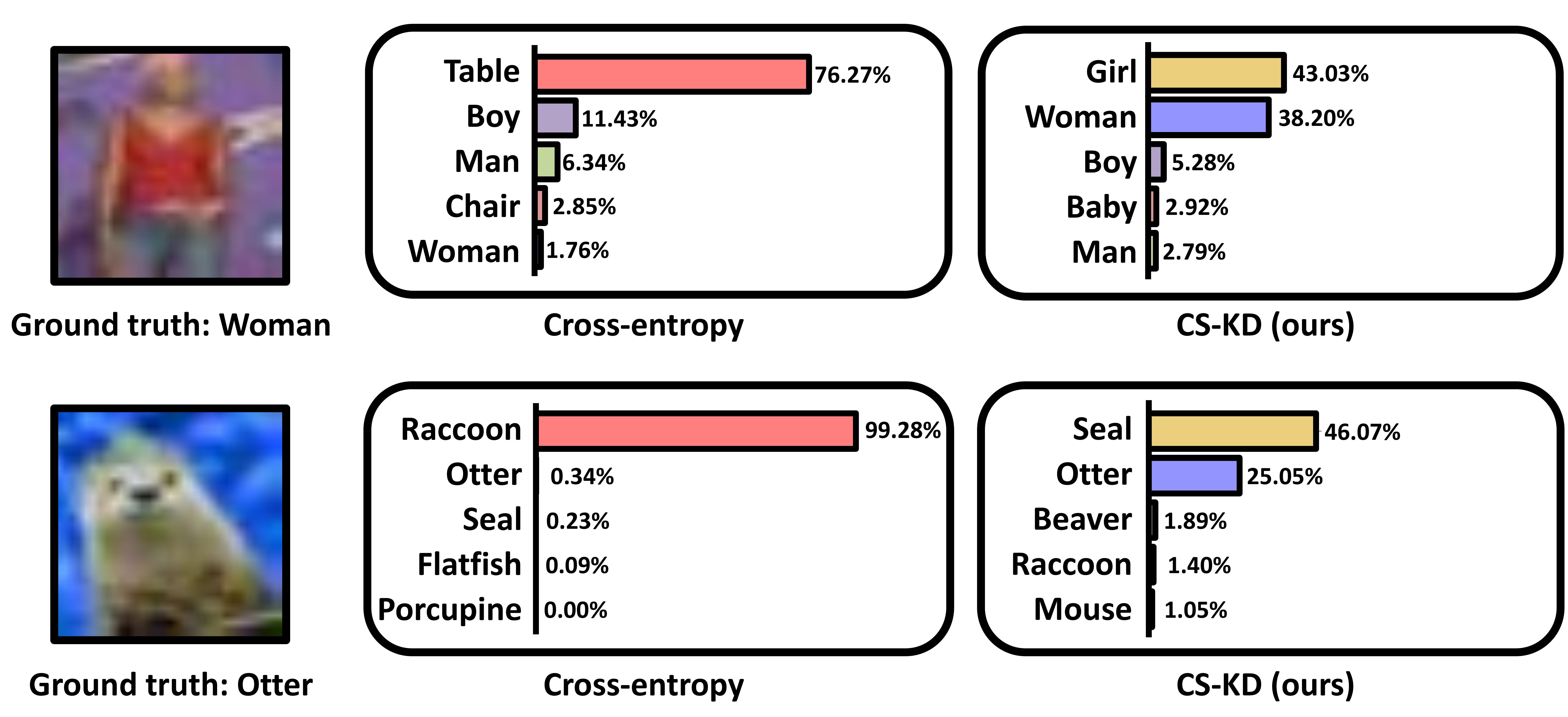}\label{fig:samples_c100}}
% \subfigure[Sample-wise regularization]
% {
% \includegraphics[width=0.48\textwidth]{figures/method_2.png}}
\caption{(a) Illustration of class-wise self-knowledge distillation (CS-KD). 
%We match the predictive distribution of DNNs between different samples with the same label.
(b) Predictive distributions on misclassified samples. We use PreAct ResNet-18 trained on CIFAR-100 dataset. 
%Networks are trained on PreAct ResNet-18 using CIFAR-100 by applying cross-entropy loss and class-wise regularization.
For misclassified samples,
softmax scores of the ground-truth class are increased by training DNNs with class-wise regularization.}
%CS-KD (ours) produces \textcolor{green}{high values of softmax scores} on the ground-truth class while cross-entropy does not.
%This implies that the class-wise regularization induces meaningful predictive distributions.}
\vspace{-0.25in}
\end{figure}

Regularizing the predictive distribution of DNNs can be effective  %address the aforementioned issues 
because it contains the most succinct knowledge of the model. % \cite{hinton2015distilling}.
On this line, 
several strategies such as {label-smoothing \cite{muller2019does, szegedy2016rethinking}}, entropy maximization~\cite{dubey2018maximum, pereyra2017regularizing},
and angular-margin based methods~\cite{chen2018virtual, zhang2019adacos} have been proposed in the literature.
They were also influential in solving related problems such as network calibration \cite{guo2017calibration}, novelty detection \cite{lee2018training},
and exploration in reinforcement learning \cite{haarnoja2018soft}.
In this paper, we focus on developing a new output regularizer for deep models utilizing the concept
of {\em dark knowledge} \cite{hinton2015distilling}, \textit{i.e.}, the knowledge on wrong predictions made by DNNs.
Its importance has been first evidenced 
by the so-called knowledge distillation (KD) \cite{hinton2015distilling}
and investigated in many following works \cite{ahn2019variational, romero2014fitnets, srinivas2018knowledge, zagoruyko2016paying}.

\begin{figure*}[t]
\vspace{-0.15in}
\centering
\subfigure[\textcolor{black}{Log-probabilities of predicted labels on misclassified samples}]
{
\includegraphics[width=0.48\textwidth]{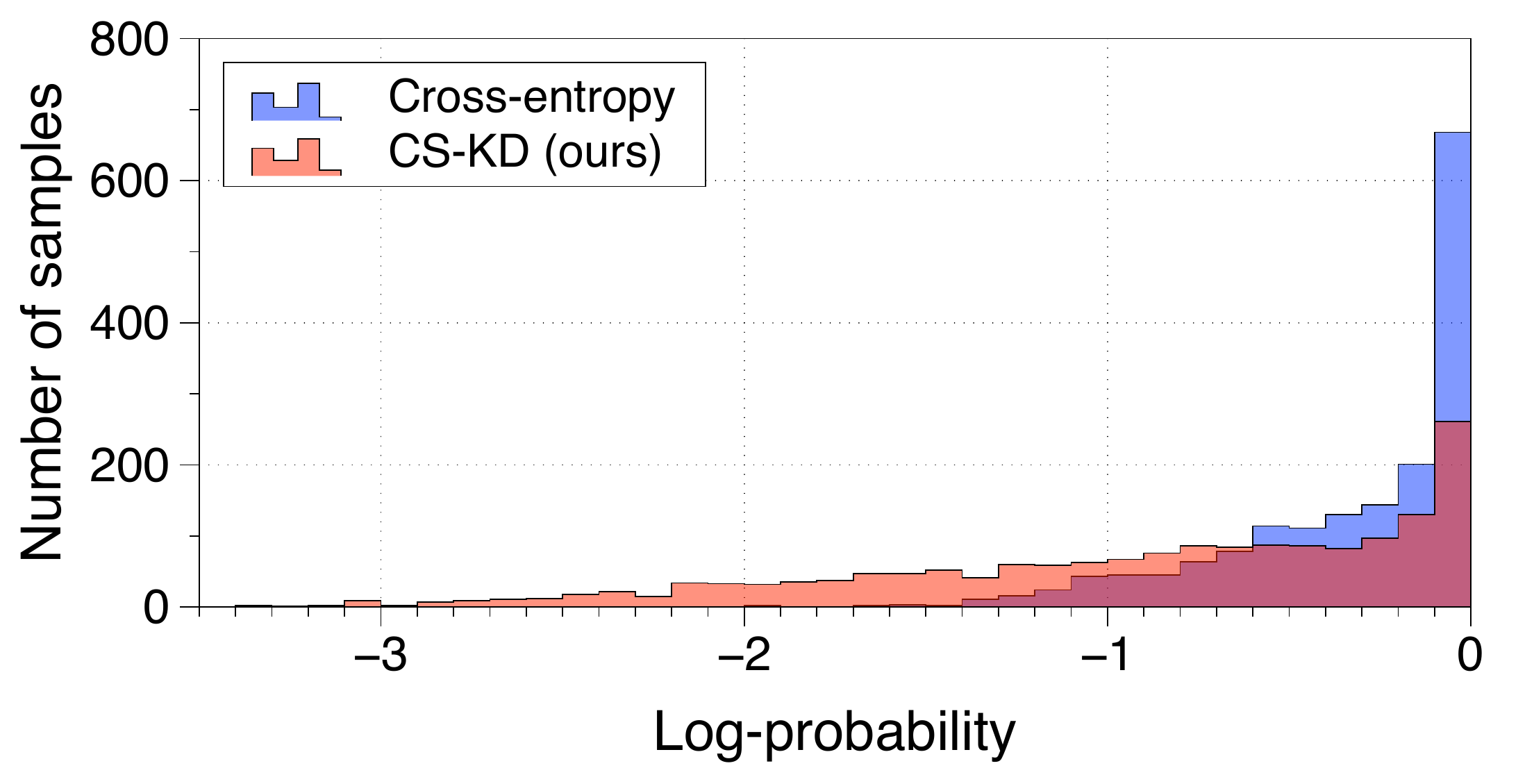}\label{fig:wrong_preds_wrong}} 
\,
\subfigure[\textcolor{black}{Log-probabilities of ground-truth labels on misclassified samples}]
{
\includegraphics[width=0.48\textwidth]{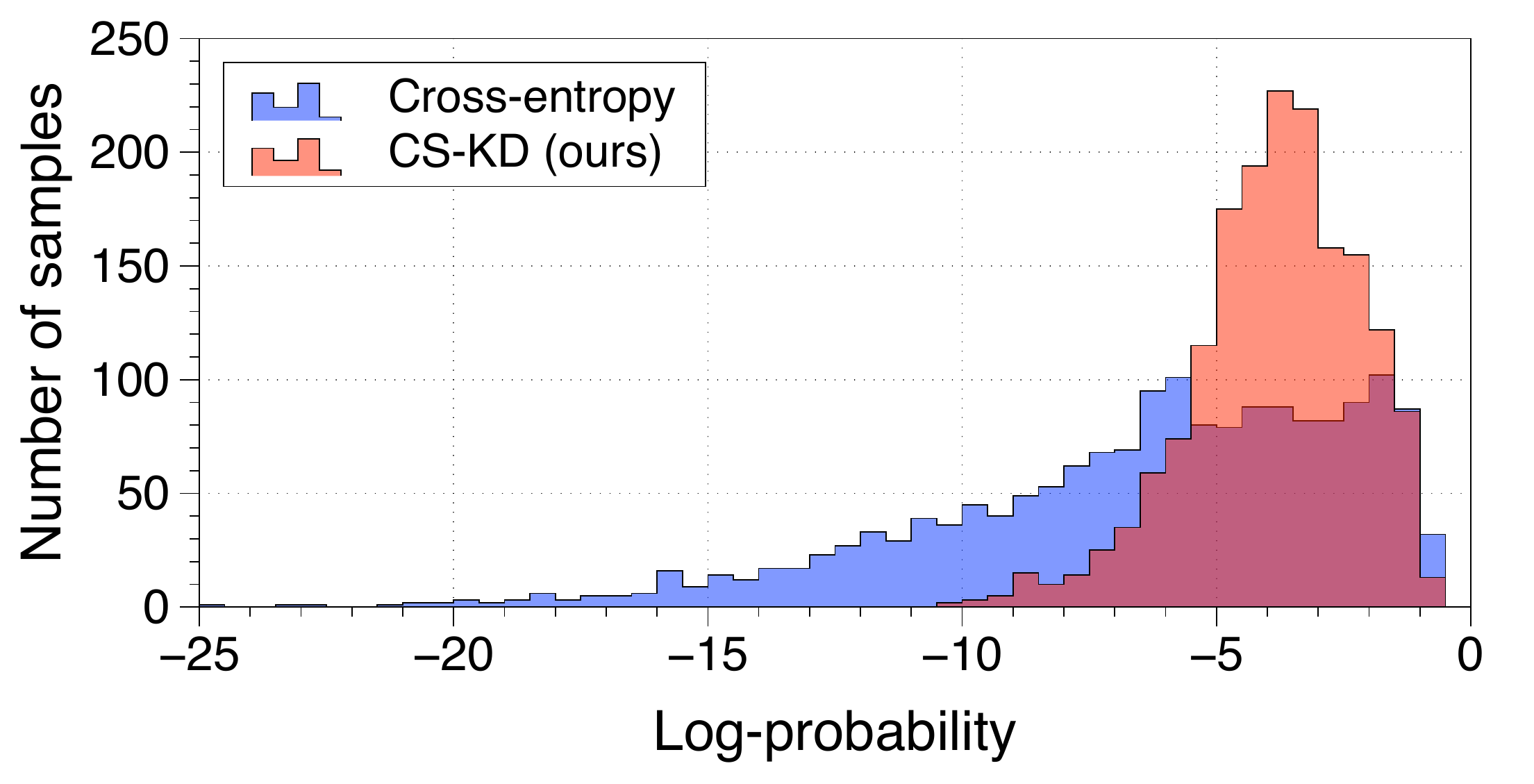}\label{fig:wrong_preds_true}}
\caption{\textcolor{black}{Histogram of log-probabilities of (a) the predicted label, \textit{i.e.}, top-1 softmax score, and (b) the ground-truth label on misclassified samples by networks trained by {the} cross-entropy {(baseline)} and CS-KD.
The networks are trained on PreAct ResNet-18 for CIFAR-100.}}
\label{fig:wrong_preds}
\vspace{-0.2in}
\end{figure*}

While the related works \cite{furlanello2018born, bagherinezhad2018label} use the knowledge distillation to transfer the dark knowledge learned by a teacher network to a student network, we regularize the dark knowledge itself during training a single network, \textit{i.e.}, self-knowledge distillation \cite{xu2019data, zhang2019your}.
Specifically,
we propose a new regularization technique, coined class-wise self-knowledge distillation (CS-KD),
that matches or distills the predictive distribution of DNNs between different samples of the same label as shown in Figure~\ref{fig:cskd}.
One can expect that the proposed regularization method forces DNNs to produce similar wrong predictions if samples are of the same class, while the conventional cross-entropy loss does not consider such consistency on the predictive distributions.
{Furthermore, it could achieve two desirable goals simultaneously: preventing overconfident predictions and reducing the intra-class variations.
We remark that they have been investigated in the literature via different methods, 
%developed as independent research, 
% \eg, 
\textit{i.e.}, entropy regularization \cite{dubey2018maximum, muller2019does, pereyra2017regularizing, szegedy2016rethinking} 
and margin-based methods \cite{chen2018virtual, zhang2019adacos}, respectively,
while we achieve both using a single principle.}
% The proposed method is the simplest way to incorporates both via a single mechanism.

We demonstrate the effectiveness of our simple yet powerful regularization method using deep convolutional neural networks,
such as ResNet~\cite{he2016deep} and DenseNet~\cite{huang2017densely} trained for image classification tasks on various datasets
including CIFAR-100~\cite{krizhevsky2009learning}, TinyImageNet\footnote{\url{https://tiny-imagenet.herokuapp.com/}}, CUB-200-2011~\cite{WahCUB_200_2011}, Stanford Dogs~\cite{KhoslaYaoJayadevaprakashFeiFei_FGVC2011},
MIT67~\cite{quattoni2009recognizing}, and 
{ImageNet \cite{deng2009imagenet}}.
%We compare or combine our method with prior regularizers.
%\cite{chen2018virtual, zhang2019adacos, dubey2018maximum, pereyra2017regularizing}.
In our experiments,
the top-1 error rates of our method are consistently lower than those of prior output regularization methods such as angular-margin based methods \cite{chen2018virtual, zhang2019adacos} and entropy regularization~{\cite{dubey2018maximum, muller2019does, pereyra2017regularizing, szegedy2016rethinking}}.
In particular, the gain tends to be larger in overall for 
the top-5 error rates and 
the expected calibration errors \cite{guo2017calibration}, which confirms that
our method indeed makes predictive distributions more meaningful.
{
We also found the top-1 error rates of our method are lower than those of the recent self-distillation methods \cite{xu2019data, zhang2019your} in overall.}
Moreover, we investigate {variants} of our method by combining it with other types of regularization methods for boosting performance, such as the Mixup regularization~\cite{zhang2017mixup} and the original KD method \cite{hinton2015distilling}. 
For example, we improve the top-1 error rate of Mixup from 37.09\% to \textcolor{black}{30.71\%}, and that of KD from 39.32\% to \textcolor{black}{34.47\%} using the CUB-200-2011 dataset under {ResNet-18 and ResNet-10}, respectively. 

We remark that the idea of using a consistency regularizer like ours
has been investigated in the literature {\cite{bachman2014learning, clark2018semi, kannan2018adversarial, miyato2018virtual, xie2019uda, tarvainen2017mean, xu2019data}}.
While most prior methods proposed to regularize the output distributions of original and 
perturbed inputs to be similar,
our method forces the consistency between different samples having the same class.
To the best of our knowledge, 
no work is known to study such a class-wise regularization.
%investigate sample-wise consistency between outputs of corrupted sample and its original sample. 
%Unlike the prior works, the proposed method regularizes class-wise consistency  
% We also investigate a variant of our method, which combines with sample-wise consistency between augmented sample and its original, 
% con in Section~\ref{sec:calibration}.
We believe that the proposed method may be influential to enjoy a broader usage in other applications, \eg, face recognition~\cite{deng2019arcface,zhang2019adacos}, and image retrieval~\cite{tolias2015particular}.
%semi-supervised learning \cite{xie2019uda} and deep reinforcement learning \cite{haarnoja2018soft}.}

\iffalse
\textcolor{blue}{Mention here or somewhere there exists many works forcing consistency in sample-wise (\eg, in semi-supervised learning), and emphasize our novelty.
We think the proposed method would enjoy a broader usage in the future (\eg, semi-supervised learning? say some concrete promising applications.}
\fi

\begin{algorithm}[h]
\caption{Class-wise self-knowledge distillation} 
\label{alg:cskd}
\begin{algorithmic}
\vspace{0.02in}
\State Initialize parameters $\theta$.
\While{$\theta$ has not converged}
    \State Sample a batch $(\mathbf{x},y)$ from the training dataset.
%    \For{($\mathbf{x},y)$ in a sampled batch}
%    \State $g_{\theta}\leftarrow 0$
    \State Sample another batch $\mathbf{x'}$ randomly, which has the same label $y$ from the training dataset.
%    \State Generate $\mathbf{x}_{\tt{aug}}$, $\mathbf{x'}_{\tt{aug}}$ using data augmentation methods.
    % \textcolor{black}{\State Update parameters $\theta$ using gradients $\nabla_{\theta}\left(\mathcal{L_{\tt{CE}}}\left(\mathbf{x}, y\right)+\lambda_{\tt{cls}}\mathcal{L_{\tt{cls}}}\left(\mathbf{x}, \mathbf{x'}\right)+\lambda_{\tt{sam}}\mathcal{L}_{\tt{sam}} \left(\mathbf{x}, \mathbf{x}_{\tt{aug}}\right)\right)$.}
    \State Update parameters $\theta$ by computing the gradients of the proposed loss function 
    $\mathcal{L}_{\tt CS{\text -}KD}(\mathbf{x}, \mathbf{x}^\prime, y; \theta, T)$ in \eqref{lam:cls}.
\EndWhile
\end{algorithmic}
\end{algorithm}

%\vspace{-0.2in}
\section{Class-wise self-knowledge distillation}\label{sec:regularizer}
%\vspace{-0.05in}
In this section, we introduce a new regularization technique named class-wise self-knowledge distillation (CS-KD).
Throughout this paper, we focus on fully-supervised classification tasks and denote $\mathbf{x}\in \mathcal{X}$ as input and $y \in \mathcal{Y}=\{1,...,C\}$ as its ground-truth label.
Suppose that a softmax classifier is used to model a posterior predictive distribution, \textit{i.e.},  
given the input $\mathbf{x}$, the predictive distribution is:
\begin{equation*}
% \vspace{-0.1in}
    P \left(y|\mathbf{x};\theta,T \right) = \frac{\exp\left(f_y\left(\mathbf{x};\theta\right)/~T\right)}{\sum_{i=1}^{C}\exp\left(f_i\left(\mathbf{x};\theta\right)/~T\right)}, 
\end{equation*}
where $f_i$ denotes the logit of DNNs for class $i$ which are parameterized by $\theta$, and $T>0$ is the temperature scaling parameter. 

% \vspace{-0.1in}
\subsection{Class-wise regularization} \label{sec:classwise}
%\vspace{-0.05in}
We consider matching the predictive distributions on samples of the same class, 
which distills their dark knowledge from the model itself. % during training a network.
To this end, we propose a class-wise regularization loss that enforces consistent predictive distributions in the same class.
Formally, given an input $\mathbf{x}$ and another randomly sampled input $\mathbf{x}^\prime$ having the same label $y$,  
it is defined as follows:
\begin{equation*}
    \mathcal{L}_{\tt{cls}}\left(\mathbf{x}, \mathbf{x}^\prime; \theta, T \right) := 
    \text{KL}\left(P(y|\mathbf{x}^\prime;\widetilde{\theta},T)
    \big\| 
    P(y|\mathbf{x};{\theta},T)\right),
\end{equation*}
where $\text{KL}$ denotes the Kullback-Leibler (KL) divergence,
and $\widetilde{\theta}$ is a fixed copy of the parameters $\theta$.
As suggested by Miyato \etal \cite{miyato2018virtual}, the gradient is not propagated through $\widetilde{\theta}$ to avoid the model collapse issue.
Similar to the original knowledge distillation method (KD; \cite{hinton2015distilling}), the proposed loss $\mathcal{L}_{\tt{cls}}$ matches two predictions. 
While the original KD matches predictions of a single sample from two networks,
we make predictions of different samples from a single network, 
\textit{i.e.}, self-knowledge distillation. %(CS-KD).
Namely, the total training loss $\mathcal{L_{\tt{\tt CS{\text -}KD}}}$ is defined as 
% a weighted sum of the two regularization terms with cross-entropy loss as 
follows:
\begin{align}
    \mathcal{L_{\tt{\tt CS{\text -}KD}}}(\mathbf{x}, \mathbf{x}^\prime, y; \theta, T) :=
      & \ \mathcal{L}_{\tt{CE}}(\mathbf{x}, y; \theta)\notag \\
    %   - y \cdot &\log P(y|\mathbf{x};{\theta},1)\notag \\
      + \ & \lambda_{\tt{cls}} \cdot T^2 \cdot \mathcal{L}_{\tt{cls}}(\mathbf{x}, \mathbf{x}^\prime; \theta, T),  \label{lam:cls}
\end{align}
% \textcolor{blue}{Check the cross entropy part of the above equation. It is wrong! And mention which term is the cross entropy.}
{where $\mathcal{L_{\tt{CE}}}$ is the standard cross-entropy loss, and }
$\lambda_{\tt{cls}}>0$ is a loss weight for the class-wise regularization. 
Note that we multiply the square of the temperature $T^2$ by following the original KD~\cite{hinton2015distilling}.
%\kimin{meaning of loss function: \eg, in other words, we consider both dark and bright knowledge from one-hot vector.}
% We also multiply square of temperature $T^2$ by following~\cite{hinton2015distilling}.
% In other words, we not only train the true label, but also regularize the wrong labels.
The full training procedure with the proposed loss $\mathcal{L}_{\tt{\tt CS{\text -}KD}}$ is summarized in Algorithm~\ref{alg:cskd}.

\subsection{Effects of class-wise regularization}
%\vspace{-0.05in}
{
The proposed CS-KD is arguably the simplest way to achieve two goals,
preventing overconfident predictions and reducing the intra-class variations,
via a single mechanism.
To avoid overconfident predictions, it utilizes the model-prediction of other samples as the soft-label.
{It is more `realistic' than the label-smoothing method \cite{muller2019does, szegedy2016rethinking}, which generates
`artificial' soft-labels.} 
%uses `artificially' generated soft-labels.}
% While the label-smoothing method \cite{muller2019does, szegedy2016rethinking} uses
% `artificially' generated soft-labels, 
% our soft-label is `realistic'.
% We hypothesis that the `realistic' labels used by our method could make a better generalization than the `artificial' ones. 
% \textcolor{blue}{KM: I don't know why this paragraph is here due to following reasons: (a) we dont compare the ours with soft-labeling in the next paragraph, (b) the goal of section is effects of proposed method not a difference with other methods. I think it is bettter to place this part at related work or experimental section..}
Besides, %$\mathcal{L}_{\tt{cls}}$ in (\ref{lam:cls}) 
ours directly minimizes the distance between two logits within the same class, and it would reduce intra-class variations.}

{
%Moreover, 
We also examined 
whether the proposed method forces
DNNs to produce meaningful predictions. To this end,
}
% To verify whether the proposed method indeed forces DNNs to produce meaningful predictions,
% \textit{i.e.}, DNNs induces similar predictive distributions on similar inputs, 
we investigate prediction values in softmax scores, \textit{i.e.}, $P(y|\mathbf{x})$, from PreAct ResNet-18 \cite{he2016identity} trained on the CIFAR-100 dataset~\cite{krizhevsky2009learning} using the standard cross-entropy loss 
and the proposed CS-KD loss.
%using CIFAR-100.
Specifically, we analyze the predictions of two concrete misclassified samples in the CIFAR-100 dataset.
As shown in Figure~\ref{fig:samples_c100},
CS-KD not only relaxes the overconfident predictions but also enhances the prediction values of classes correlated to the ground-truth class.
%prediction values in softmax scores of the ground-truth classes. 
%The aspect of the high values in softmax scores of the ground-truth classes
This implies that CS-KD induces meaningful predictions by forcing DNNs to produce similar predictions on similar inputs.
To evaluate the prediction quality,
we also report log-probabilities of the softmax scores on the predicted class and ground-truth class on samples that are commonly misclassified by both the cross-entropy and our method.
As shown in Figure~\ref{fig:wrong_preds_wrong}, 
our method produces less confident predictions on misclassified samples compared to the cross-entropy method.
Interestingly, our method increases ground-truth scores for misclassified samples, as reported in Figure \ref{fig:wrong_preds_true}. 
{In our experiments, we found that the classification accuracy and calibration effects can be improved by forcing DNNs to produce such meaningful predictions (see Section~\ref{sec:classification} and \ref{sec:calibration}).}

\begin{table*}[t]
\begin{center}
\resizebox{\textwidth}{!}{
\begin{tabular}{ccccccc}
\toprule
Model & Method & CIFAR-100 & TinyImageNet & CUB-200-2011 & Stanford Dogs & MIT67
\\ \midrule
\multirow{6}{*}{ResNet-18} 
& Cross-entropy\hfill \,   & {{24.71}\ms{0.24}\hfill} \, & {{43.53}\ms{0.19}\hfill} \, & {{46.00}\ms{1.43}\hfill} \, & {{36.29}\ms{0.32}\hfill} \, & {{44.75}\ms{0.80}\hfill} \,     \\
& AdaCos\hfill \,          & {{23.71}\ms{0.36}\hfill} \, & {{42.61}\ms{0.20}\hfill} \, & {{35.47}\ms{0.07}\hfill} \, & {{32.66}\ms{0.34}\hfill} \, & {{42.66}\ms{0.43}\hfill} \,     \\
& Virtual-softmax\hfill \, & {{23.01}\ms{0.42}\hfill} \, & {{42.41}\ms{0.20}\hfill} \, & {{35.03}\ms{0.51}\hfill} \, & {{31.48}\ms{0.16}\hfill} \, & {{42.86}\ms{0.71}\hfill} \,     \\
& Maximum-entropy\hfill \, & {{22.72}\ms{0.29}\hfill} \, & {{41.77}\ms{0.13}\hfill} \, & {{39.86}\ms{1.11}\hfill} \, & {{32.41}\ms{0.20}\hfill} \, & {{43.36}\ms{1.62}\hfill} \,     \\
& {Label-smoothing}\hfill \, & {{22.69}\ms{0.28}\hfill} \, & {{43.09}\ms{0.34}\hfill} \, & {{42.99}\ms{0.99}\hfill} \, & {{35.30}\ms{0.66}\hfill} \, & {{44.40}\ms{0.71}\hfill} \,     \\
% & \textcolor{red}{Clean Logit Pairing}\hfill \, & {{22.69}\ms{0.40}\hfill} \, & {{42.51}\ms{0.33}\hfill} \, & {{36.45}\ms{0.69}\hfill} \, & {{32.26}\ms{0.35}\hfill} \, & {{41.69}\ms{1.25}\hfill} \,     \\
& CS-KD (ours)\hfill \,    & \textbf{{21.99}\ms{0.13}\hfill} \, \textbf{(-11.0\%)}& \textbf{{41.62}\ms{0.38}\hfill} \, \textbf{(- 4.4\%)}& \textbf{{33.28}\ms{0.99}\hfill} \, \textbf{(-27.7\%)}& \textbf{{30.85}\ms{0.28}\hfill} \, \textbf{(-15.0\%)}& \textbf{{40.45}\ms{0.45}\hfill} \, \textbf{(- 9.6\%)}    \\
\midrule \multirow{6}{*}{DeseNet-121} 
& Cross-entropy\hfill \,   & {{22.23}\ms{0.04}\hfill} \, & {{39.22}\ms{0.27}\hfill} \, & {{42.30}\ms{0.44}\hfill} \, & {{33.39}\ms{0.17}\hfill} \, & {{41.79}\ms{0.19}\hfill} \,     \\
& AdaCos\hfill \,          & {{22.17}\ms{0.24}\hfill} \, & {{38.76}\ms{0.23}\hfill} \, & {{30.84}\ms{0.38}\hfill} \, & {{27.87}\ms{0.65}\hfill} \, & {{40.25}\ms{0.68}\hfill} \,     \\
& Virtual-softmax\hfill \, & {{23.66}\ms{0.10}\hfill} \, & {{41.58}\ms{1.58}\hfill} \, & {{33.85}\ms{0.75}\hfill} \, & {{30.55}\ms{0.72}\hfill} \, & {{43.66}\ms{0.30}\hfill} \,     \\
& Maximum-entropy\hfill \, & {{22.87}\ms{0.45}\hfill} \, & {{38.39}\ms{0.33}\hfill} \, & {{37.51}\ms{0.71}\hfill} \, & {{29.52}\ms{0.74}\hfill} \, & {{43.48}\ms{1.30}\hfill} \,     \\
& {Label-smoothing}\hfill \, & {{21.88}\ms{0.45}\hfill} \, & {{38.75}\ms{0.18}\hfill} \, & {{40.63}\ms{0.24}\hfill} \, & {{31.39}\ms{0.46}\hfill} \, & {{42.24}\ms{1.23}\hfill} \,     \\
% & \textcolor{red}{Clean Logit Pairing}\hfill \, & {{21.83}\ms{0.44}\hfill} \, & {{39.09}\ms{0.80}\hfill} \, & {{32.22}\ms{0.50}\hfill} \, & {{29.50}\ms{0.58}\hfill} \, & {{42.64}\ms{0.63}\hfill} \,     \\
& CS-KD (ours)\hfill \,    & \textbf{{21.69}\ms{0.49}\hfill} \, \textbf{(- 2.4\%)}& \textbf{{37.96}\ms{0.09}\hfill} \, \textbf{(- 3.2\%)}& \textbf{{30.83}\ms{0.39}\hfill} \, \textbf{(-27.1\%)}& \textbf{{27.81}\ms{0.13}\hfill} \, \textbf{(-16.7\%)}& \textbf{{40.02}\ms{0.91}\hfill} \, \textbf{(- 4.2\%)}    \\
\bottomrule
\end{tabular}}
\end{center}
\caption{Top-1 error rates (\%) on various image classification tasks and model architectures. 
 We report the mean and standard deviation over three runs with different random seeds. Values in parentheses indicate relative error rate reductions from the cross-entropy,
 and the best results are indicated in bold.
}\label{tbl:main}
\vspace{-0.05in}
\end{table*}

\begin{table*}[h]
\begin{center}
\resizebox{\textwidth}{!}{
\begin{tabular}{cccccc}
\toprule
Method & CIFAR-100 & TinyImageNet & CUB-200-2011 & Stanford Dogs & MIT67
\\ \midrule
Cross-entropy   & {{24.71}\ms{0.24}\hfill} \, & {{43.53}\ms{0.19}\hfill} \, & {{46.00}\ms{1.43}\hfill} \, & {{36.29}\ms{0.32}\hfill} \, & {{44.75}\ms{0.80}\hfill} \,     \\
DDGSD & {{23.85}\ms{1.57}\hfill} \, & \textbf{{41.48}\ms{0.12}\hfill} \, & {{41.17}\ms{1.28}\hfill} \, & {{31.53}\ms{0.54}\hfill} \, & {{41.17}\ms{2.46}\hfill} \,     \\
BYOT & {{23.81}\ms{0.11}\hfill} \, & {{44.02}\ms{0.57}\hfill} \, & {{40.76}\ms{0.39}\hfill} \, & {{34.02}\ms{0.14}\hfill} \, & {{44.88}\ms{0.46}\hfill} \,     \\
CS-KD (ours)    & \textbf{{21.99}\ms{0.13}\hfill} \, \textbf{(-11.0\%)}& {{41.62}\ms{0.38}\hfill} \, {(- 4.4\%)}& \textbf{{33.28}\ms{0.99}\hfill} \, \textbf{(-27.7\%)}& \textbf{{30.85}\ms{0.28}\hfill} \, \textbf{(-15.0\%)}& \textbf{{40.45}\ms{0.45}\hfill} \, \textbf{(- 9.6\%)}    \\
\bottomrule
\end{tabular}}
\end{center}
\caption{{Top-1 error rates (\%) of ResNet-18 with self-distillation methods on various image classification tasks. 
 We report the mean and standard deviation over three runs with different random seeds. Values in parentheses indicate relative error rate reductions from the cross-entropy, and
 the best results are indicated in bold. The self-distillation methods are re-implemented under our code-base.}
}\label{tbl:self}
\vspace{-0.2in}
\end{table*}

% \vspace{-0.05in}
\section{Experiments}

\subsection{Experimental setup} \label{exp:setup} %\sm{write only common thing, others write indiviually}

% \vspace{0.02in}
\noindent\textbf{Datasets.} 
To demonstrate our method under general situations of data diversity, 
we consider various %image datasets
image classification tasks, including conventional classification and fine-grained classification tasks.\footnote{Code is available at \url{https://github.com/alinlab/cs-kd}.}
Specifically, we use CIFAR-100~\cite{krizhevsky2009learning} and TinyImageNet\footnote{\url{https://tiny-imagenet.herokuapp.com/}} datasets for conventional classification tasks,
and CUB-200-2011~\cite{WahCUB_200_2011}, Stanford Dogs \cite{KhoslaYaoJayadevaprakashFeiFei_FGVC2011}, and MIT67 \cite{quattoni2009recognizing} datasets for fine-grained classification tasks. 
The fine-grained image classification tasks have visually similar classes and consist of fewer training samples per class compared to conventional classification tasks.
{ImageNet \cite{deng2009imagenet} is used for a large-scale classification task.}
% We sample 10\% of the training dataset randomly as a validation set \textcolor{black}{for TinyImageNet} and report the test accuracy based on the validation accuracy.}
% For the other datasets, we report the best validation accuracy.

\vspace{0.05in}
\noindent\textbf{Network architecture.} 
We consider two state-of-the-art convolutional neural network architectures:
ResNet \cite{he2016deep} and DenseNet \cite{huang2017densely}.
We use standard ResNet-18 with 64 filters and DenseNet-121 with a growth rate of 32 for image size $224 \times 224$. 
For CIFAR-100 and TinyImageNet, \textcolor{black}{we use PreAct ResNet-18 \cite{he2016identity}, which modifies} the first convolutional layer\footnote{We used a reference implementation: \url{https://github.com/kuangliu/pytorch-cifar}.} 
with kernel size $3 \times 3$, strides 1 and padding 1,
instead of the kernel size $7 \times 7$, strides 2 and padding 3, for image size $32 \times 32$ by following \cite{zhang2017mixup}. We use DenseNet-BC structure~\cite{huang2017densely}, and the first convolution layer of the network is also modified in the same way as in PreAct ResNet-18 for image size $32 \times 32$.

\vspace{0.05in}
\noindent\textbf{Hyper-parameters.}
All networks are trained from scratch and optimized by stochastic gradient descent (SGD) with momentum 0.9, weight decay 0.0001, and an initial learning rate of 0.1. The learning rate is divided by 10 after epochs 100 and 150 for all datasets, and total epochs are 200. We set batch size as 128 for conventional, and 32 for fine-grained classification tasks. 
{We use the standard data augmentation technique for ImageNet \cite{deng2009imagenet}, \textit{i.e.}, flipping and random cropping.}
% we use standard augmentation technique used for \textcolor{blue}{ImageNet \cite{deng2009imagenet} (JW: cite!)} (\textit{i.e.}, flipping and random sized cropping). 
For our method,
the temperature $T$ is chosen from $\{1, 4\}$, and the loss weight $\lambda_{\tt cls}$ is chosen from $\{1,2,3,4\}$.
The optimal parameters are chosen to minimize the top-1 error rates on the validation set. 
More detailed ablation studies on the hyper-parameters 
$T$ and $\lambda_{\tt cls}$ are provided in the supplementary material.

\vspace{0.05in}
\noindent\textbf{Baselines.} 
We compare our method with prior regularization methods %\textcolor{orange}{which modify predictive distribution}%for output regularizers 
such as the state-of-the-art angular-margin based methods \cite{chen2018virtual,zhang2019adacos}, entropy regularization \cite{dubey2018maximum, muller2019does, pereyra2017regularizing, szegedy2016rethinking}
and self-distillation methods \cite{xu2019data,zhang2019your}.
They also regularize predictive distributions like ours. \vspace{-0.05in}
% \textbf{ - AdaCos} \cite{zhang2019adacos}.\footnote{We used a reference implementation: \url{https://github.com/4uiiurz1/pytorch-adacos}} 
% AdaCos dynamically scales the cosine similarities between training samples and corresponding class center vectors to maximize angular-margin. \\
% \textbf{ - Virtual-softmax} \cite{chen2018virtual}. Virtual-softmax injects an additional virtual class to maximize angular-margin. \\
% \textbf{ - Maximum-entropy} \cite{dubey2018maximum, pereyra2017regularizing}. Maximum-entropy is a typical entropy regularization, which maximizes the entropy of the predictive distribution. \\
% \textcolor{red}{
% \textbf{ - Label-smoothing} \cite{muller2019does, szegedy2016rethinking}. Label-smoothing uses soft labels that are a weighted average of the one-hot labels and the uniform distribution. \\
% \textbf{ - Data-distortion Self-distillation} \cite{xu2019data}. 
% Data-distortion Self-distillation is one of the consistency regularization techniques, which forces the consistent outputs across different augmented versions of the data.\\
% \textbf{ - Teacher-student Self-distillation} \cite{zhang2019your}. 
% Teacher-student Self-distillation transfers the knowledge in the
% deeper portion of the networks into the shallow ones. \\}
\begin{itemize}
    \item \textbf{AdaCos} \cite{zhang2019adacos}.\footnote{We used a reference implementation: \url{https://github.com/4uiiurz1/pytorch-adacos}} 
    AdaCos dynamically scales the cosine similarities between training samples and corresponding class center vectors to maximize angular-margin.\vspace{-0.1in}
    \item \textbf{Virtual-softmax} \cite{chen2018virtual}. Virtual-softmax injects an additional virtual class to maximize angular-margin.\vspace{-0.1in}
    \item \textbf{Maximum-entropy} \cite{dubey2018maximum, pereyra2017regularizing}. Maximum-entropy is a typical entropy regularization, which maximizes the entropy of the predictive distribution.\vspace{-0.1in}
    {\item \textbf{Label-smoothing} \cite{muller2019does, szegedy2016rethinking}. Label-smoothing uses soft labels that are a weighted average of the one-hot labels and the uniform distribution.\vspace{-0.1in}
    \item \textbf{DDGSD} \cite{xu2019data}. Data-distortion guided self-distillation (DDGSD) is one of the consistency regularization techniques, which forces the consistent outputs across different augmented versions of the data.\vspace{-0.1in}
    \item \textbf{BYOT} \cite{zhang2019your}. Be Your Own Teacher (BYOT) transfers the knowledge in the deeper portion of the networks into the shallow ones. \vspace{-0.1in}}
\end{itemize}

\vspace{0.1in}
\noindent\textbf{Evaluation metric.}\label{eval_metric}
For evaluation, we measure the following metrics:
\begin{itemize}
\item \textbf{Top-1~/~5 error rate.} The top-$k$ error rate is the fraction of %misclassification within 
test samples for which the correct label is not in the top-$k$ confidences.
We measure top-1 and top-5 error rates to evaluate the generalization performances.
\vspace{-0.2in}
\item\textbf{Expected Calibration Error (ECE).} ECE~\cite{guo2017calibration, naeini2015obtaining} approximates the difference in
expectation between confidence and accuracy. It is calculated by partitioning predictions into $M$ equally-spaced bins and taking a weighted average of bins' difference of confidence and accuracy, \textit{i.e.}, $\text{ECE} =\sum_{m=1}^M \frac{|B_m|}{n} |\mathrm{acc}(B_m) - \mathrm{conf}(B_m)|$,
where $n$ is the number of samples, $B_m$ is the set of samples whose confidence falls into the $m$-th interval, and $\mathrm{acc}(B_m)$, $\mathrm{conf}(B_m)$ are the accuracy and the average confidence of $B_m$, respectively. We measure ECE with 20 bins
% , \textit{i.e.}, $M=20$, 
to evaluate whether the model represents the true correctness likelihood. 
\vspace{-0.1in}
% \kimin{acc / conf}
%\centerline{\textbf{ECE} $=\sum_{m=1}^M \frac{|B_m|}{n} |\mathrm{acc}(B_m) - \mathrm{conf}(B_m)|,$ where $n$ is the number of samples.} 
\item\textbf{Recall at $\boldsymbol{k}$ (\textbf{R@$\boldsymbol{k}$}).} Recall at $k$ is the percentage of test samples that have at least one from the same class in $k$ nearest neighbors on the feature space. To measure the distance between two samples, we use $L_2$-distance between their pooled features of the penultimate layer.
%$L_2$-distance of average-pooled features of the penultimate layer is used as a distance metric between two samples. 
{We compare the recall at $k=1$ scores to evaluate intra-class variations of learned features.}\vspace{-0.1in}
\end{itemize}

\begin{table*}[t]
\begin{center}
\begin{tabular}{cccccc}
\toprule
Method           & CIFAR-100                       & TinyImageNet                  & CUB-200-2011                          & Stanford Dogs                   & MIT67 \\ \midrule
Cross-entropy    & \textcolor{black}{{24.71}\ms{0.24}}          & {43.53}\ms{0.19}          & {46.00}\ms{1.43}          & {36.29}\ms{0.32}          & {44.75}\ms{0.80}  \\
CS-KD (ours)     & \textcolor{black}{{21.99}\ms{0.13}}          & {41.62}\ms{0.38}          & \textcolor{black}{{33.28}\ms{0.99}}          & \textcolor{black}{{30.85}\ms{0.28}}          & \textcolor{black}{{40.45}\ms{0.45}}  \\
Mixup            & \textcolor{black}{{21.67}\ms{0.34}}          & {41.57}\ms{0.38}          & {37.09}\ms{0.27}          & {32.54}\ms{0.04}                   & {41.67}\ms{1.05}  \\
Mixup + CS-KD (ours)   & \textcolor{black}{\textbf{{20.40}\ms{0.31}}}          & \textbf{{40.71}\ms{0.32}}          & \textcolor{black}{\textbf{{30.71}\ms{0.64}}}          & \textcolor{black}{\textbf{{29.93}\ms{0.14}}}         & \textcolor{black}{\textbf{{39.65}\ms{0.85}}}  \\
\bottomrule
\end{tabular}
\end{center}
\caption{Top-1 error rates (\%) of ResNet-18 with Mixup regularization on various image classification tasks.
We report the mean and standard deviation over three runs with different random seeds, and the best results are indicated in bold.}\label{tbl:mixup}
\vspace{-0.05in}
\end{table*}

\begin{table*}[t]
\begin{center}
\begin{tabular}{cccccc}
\toprule
Method              & CIFAR-100 & TinyImageNet & CUB-200-2011 & Stanford Dogs & MIT67 \\ \midrule
Cross-entropy       & \textcolor{black}{{26.72}\ms{0.33}}             & {46.61}\ms{0.22}             & {48.36}\ms{0.61}             & {38.96}\ms{0.40}             & {44.75}\ms{0.62} \\
CS-KD (ours)        & \textcolor{black}{{25.80}\ms{0.10}}             & {44.67}\ms{0.12}             & {39.12}\ms{0.09}             & {34.07}\ms{0.46}             & {41.54}\ms{0.67} \\
KD                  & \textcolor{black}{{25.84}\ms{0.07}}             & {43.31}\ms{0.11}             & {39.32}\ms{0.65}             & {34.23}\ms{0.42}             & \textcolor{black}{{41.47}\ms{0.79}} \\
KD + CS-KD (ours)   & \textcolor{black}{\textbf{{25.58}\ms{0.16}}}    & \textbf{{42.82}\ms{0.33}}    & \textcolor{black}{\textbf{{34.47}\ms{0.17}}}    & \textcolor{black}{\textbf{{32.59}\ms{0.50}}}    & \textcolor{black}{\textbf{{40.27}\ms{0.78}}} \\
\bottomrule
\end{tabular}
\end{center}
\caption{Top-1 error rates (\%) of ResNet-10 (student) with knowledge distillation (KD) on various image classification tasks. Teacher networks are pre-trained on DenseNet-121 by CS-KD.
We report the mean and standard deviation over three runs with different random seeds, and the best results are indicated in bold.}\label{tbl:kd}
\vspace{-0.2in}
\end{table*}

\begin{table}[b]
\vspace{-0.2in}
\begin{center}
\scalebox{0.95}{
\begin{tabular}{ccccc} 
\toprule
Model & Method & Top-1 (1-crop) \\ 
\midrule
\multirow{2}{*}{ResNet-50} 
& Cross-entropy            & 24.0  \\
& CS-KD (ours)             & \textbf{23.6}  \\
\midrule
\multirow{2}{*}{ResNet-101} 
& Cross-entropy            & 22.4  \\
& CS-KD (ours)             & \textbf{22.0}  \\
\midrule
\multirow{2}{*}{ResNeXt-101-32x4d}
& Cross-entropy            & 21.6  \\
& CS-KD (ours)             & \textbf{21.2}  \\
\bottomrule
\end{tabular}}
\end{center}
% \vspace{0.1in}
\caption{
Top-1 error rates (\%) on ImageNet dataset with various model architectures trained for 90 epochs with batch size 256.
The best results are indicated in bold.}\label{tbl:imagenet}
\end{table}

\begin{figure*}[h] \vspace{-0.2in}
\centering
\subfigure[{Cross-entropy}]
{
\includegraphics[width=0.23\textwidth]{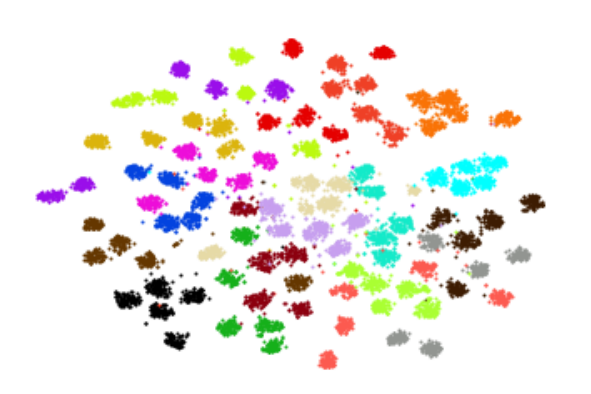}\label{fig:tsne_ce}} 
\,
\subfigure[{Virtual-softmax}]
{
\includegraphics[width=0.23\textwidth]{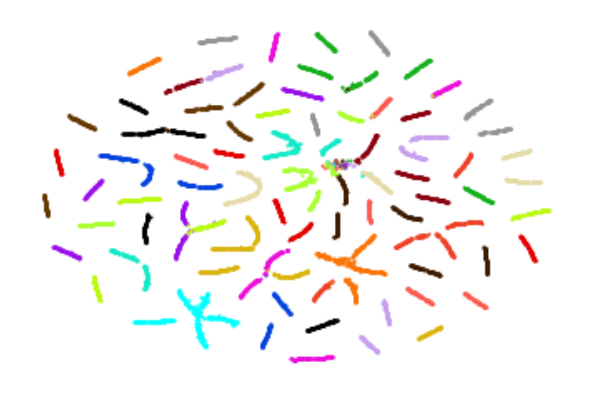}\label{fig:tsne_vs}}
\,
\subfigure[{AdaCos}]
{
\includegraphics[width=0.23\textwidth]{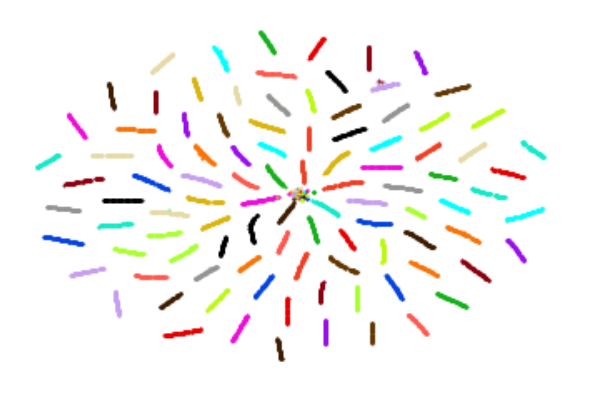}\label{fig:tsne_adacos}}
\,
% \subfigure[{Maximum-entropy}]
% {
% \includegraphics[width=0.31\textwidth]{figures/tSNE_c100_me.pdf}\label{fig:tsne_me}} 
% \,
% \subfigure[{Label-smoothing}]
% {
% \includegraphics[width=0.31\textwidth]{figures/tSNE_c100_ls.pdf}\label{fig:tsne_mx}}
% \,
\subfigure[{CS-KD (ours)}] 
{
\includegraphics[width=0.23\textwidth]{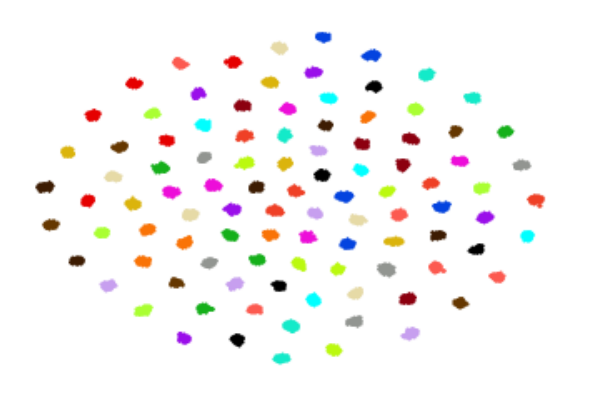}  \label{fig:tsne_cskd}}
% \,
% \subfigure[Mixup + CS-KD (ours)]
% {
% \includegraphics[width=0.3\textwidth]{figures/tSNE_c100_mxcskd.pdf} }
\vspace{0.05in}
\caption{
Visualization of various feature embeddings on the penultimate layer using t-SNE on PreAct ResNet-18 for CIFAR-100.
% (a) Cross-entropy, (b) Virtual-softmax, (c) AdaCos, (d) Maximum-entropy, (e) Mixup and (f) CS-KD.
% from 10,000 number of randomly chosen training samples of CIFAR-100. 
% We found that our method reduces that the intra-class variartions significantly. 
% Note that 20 superclasses in CIFAR-100 are drawn by 20 different colors. 
The proposed method (d) shows the smallest intra-class variation that leads to the best top-1 error rate.
}\vspace{-0.1in}
%Best viewed in color.
\label{fig:tsne_all}
\end{figure*}

\begin{figure*}[t] \vspace{-0.05in}
\centering
\subfigure[\textcolor{black}{Cross-entropy}]
{\includegraphics[width=0.3\textwidth]{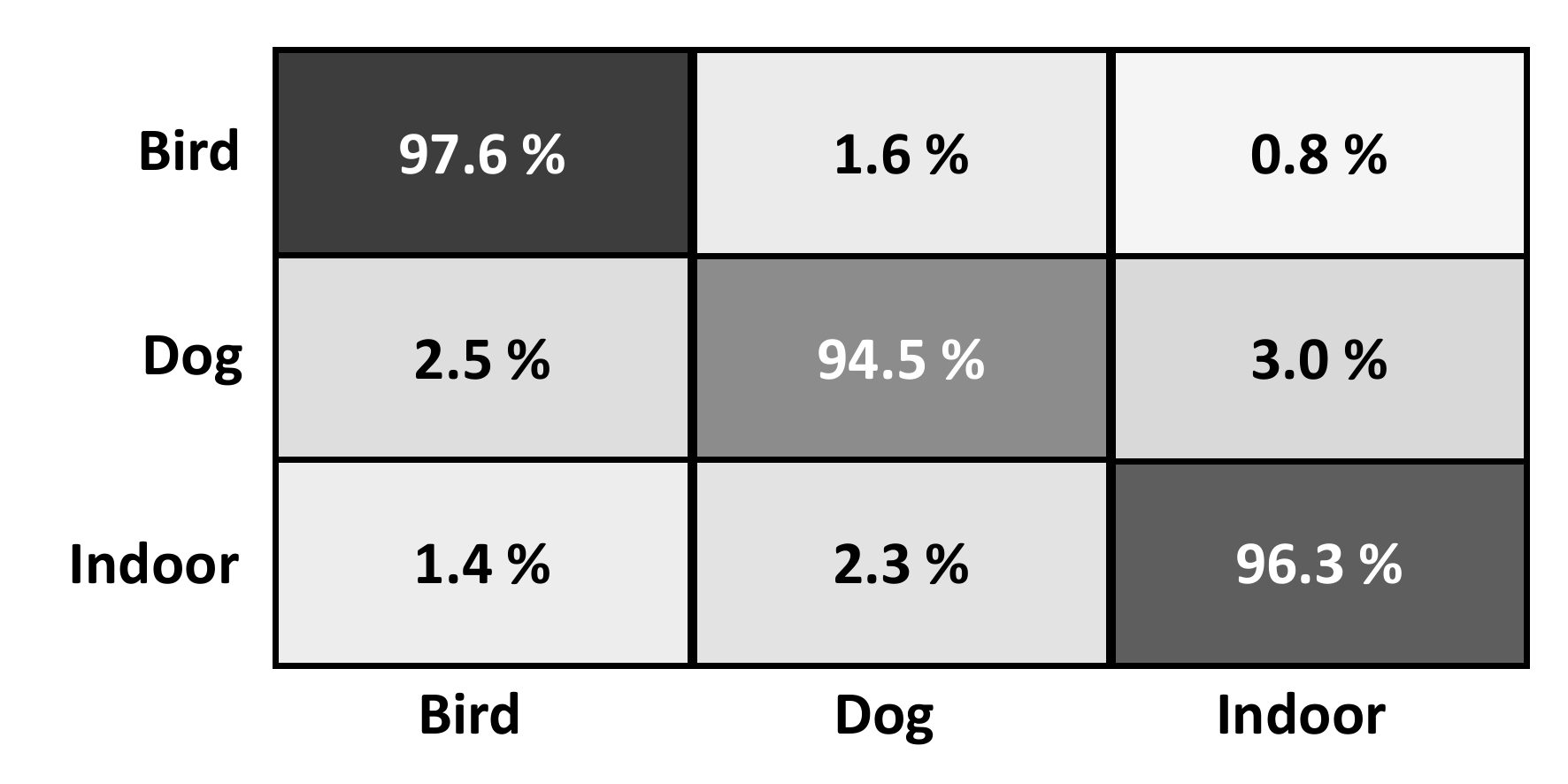}\label{fig:matrix:ce}} 
\,
\subfigure[\textcolor{black}{CS-KD (ours)}]
{
\includegraphics[width=0.3\textwidth]{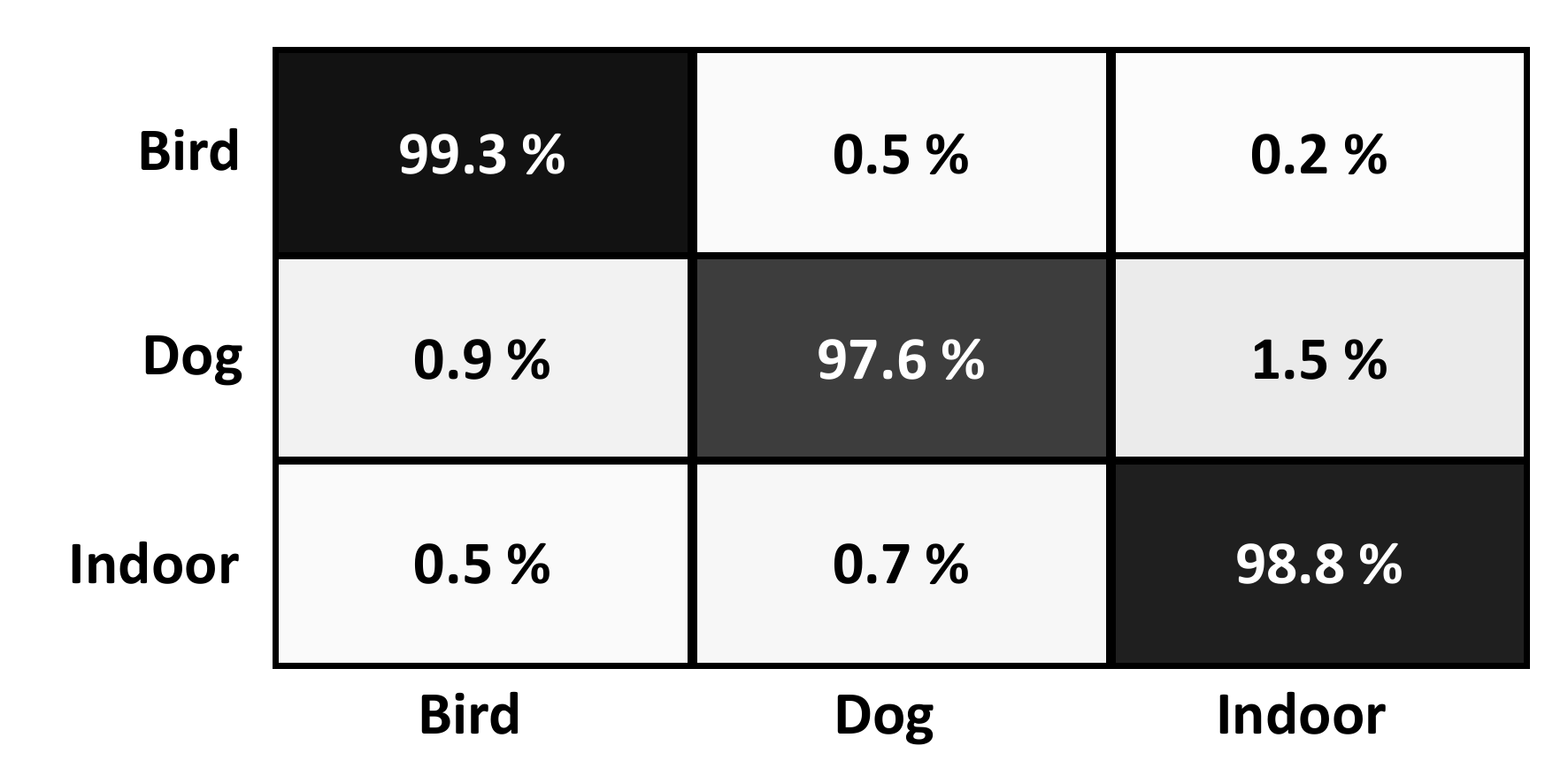}\label{fig:matrix:kl}}
\,
\subfigure[\textcolor{black}{Top-1 error rates (\%)}]
{
\includegraphics[width=0.3\textwidth]{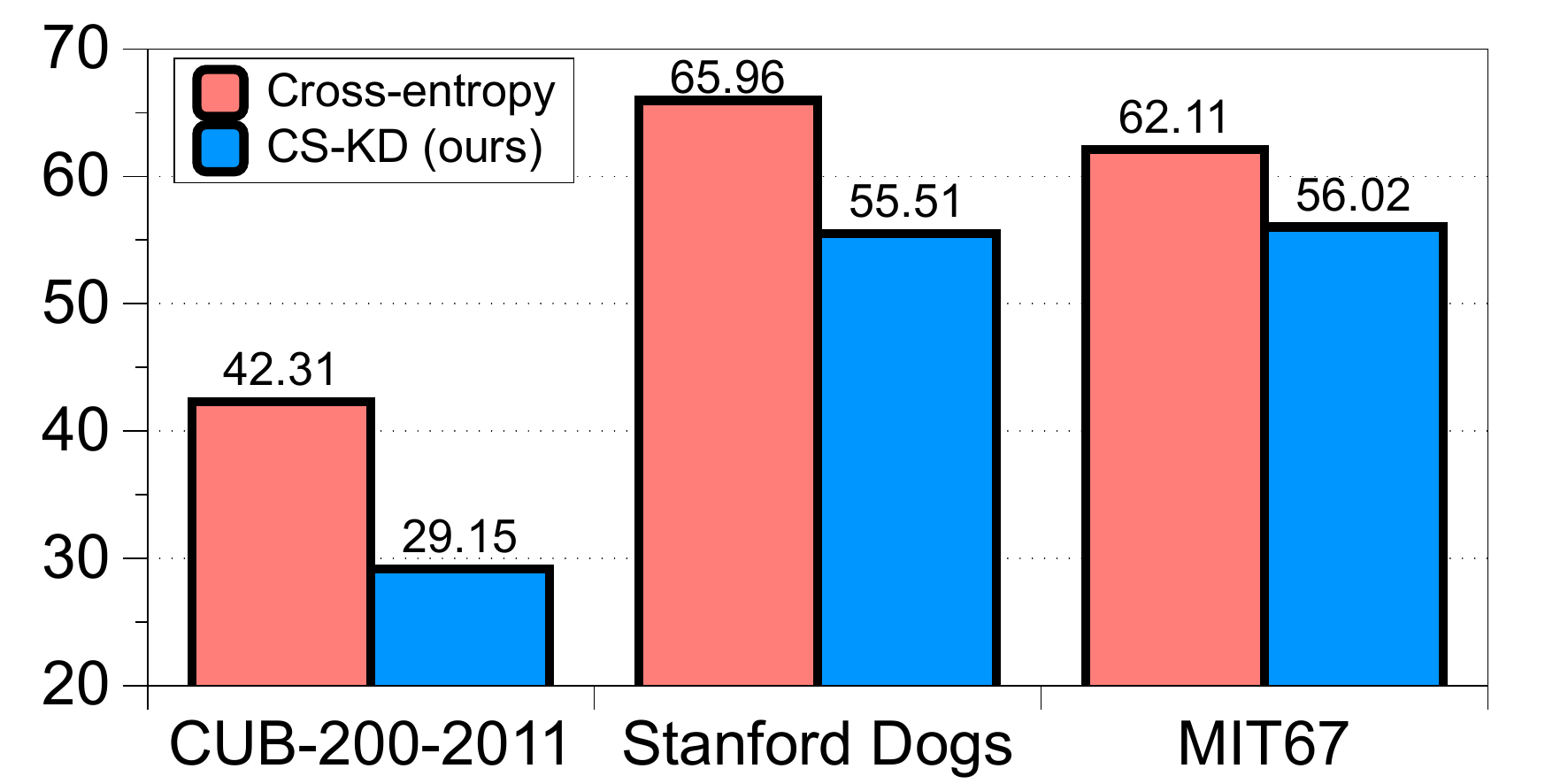}\label{tbl:conf}}
\vspace{0.05in}
\caption{Experimental results of ResNet-18 on the mixed dataset. 
The hierarchical classification accuracy (\%) of each model trained by (a) the cross-entropy and (b) our method. One can observe that the model
trained by CS-KD is less confusing classes across different domains. (c) Top-1 error rates (\%) of fine-grained label classification.}
\label{fig:confusion-matrix}
\vspace{-0.2in}
\end{figure*}

\begin{table*}[t]
\vspace{-0.15in}
\begin{center}
\resizebox{\textwidth}{!}{
\begin{tabular}{ccccccc}
\toprule
Measurement & Method & CIFAR-100 & TinyImageNet & CUB-200-2011 & Stanford Dogs & MIT67
\\ \midrule
\multirow{7}{*}{Top-5 $\downarrow$} 
& Cross-entropy\hfill \,   & {{6.91}\ms{0.09}\hfill} \, & {{22.21}\ms{0.29}\hfill} \, & {{22.30}\ms{0.68}\hfill} \, & {{11.80}\ms{0.27}\hfill} \, & {{19.25}\ms{0.53}\hfill} \,     \\
& AdaCos\hfill \,         & {{9.99}\ms{0.20}\hfill} \, & {{22.24}\ms{0.11}\hfill} \, & {{15.24}\ms{0.66}\hfill} \, & {{11.02}\ms{0.22}\hfill} \, & {{19.05}\ms{2.33}\hfill} \,     \\
& Virtual-softmax\hfill \, & {{8.54}\ms{0.11}\hfill} \, & {{24.15}\ms{0.17}\hfill} \, & {{13.16}\ms{0.20}\hfill} \, & {{8.64}\ms{0.21}\hfill} \, & {{19.10}\ms{0.20}\hfill} \,     \\
& Maximum-entropy\hfill \, & {{7.29}\ms{0.12}\hfill} \, & {{21.53}\ms{0.50}\hfill} \, & {{19.80}\ms{1.21}\hfill} \, & {{10.90}\ms{0.31}\hfill} \, & {{20.47}\ms{0.90}\hfill} \,     \\
& {Label-smoothing}\hfill \, & {{7.18}\ms{0.08}\hfill} \, & {{20.74}\ms{0.31}\hfill} \, & {{22.40}\ms{0.85}\hfill} \, & {{13.41}\ms{0.40}\hfill} \, & {{19.53}\ms{0.75}\hfill} \,     \\
% & \textcolor{red}{Clean Logit Pairing}\hfill \, & {{7.77}\ms{0.12}\hfill} \, & {{21.84}\ms{0.99}\hfill} \, & {{17.54}\ms{0.16}\hfill} \, & {{12.01}\ms{0.71}\hfill} \, & {{19.88}\ms{0.90}\hfill} \,     \\
& CS-KD (ours)\hfill \,    & \textbf{{5.69}\ms{0.03}}\hfill \, & {{19.21}\ms{0.04}\hfill} \, & \textbf{{13.07}\ms{0.26}}\hfill \, & \textbf{{8.55}\ms{0.07}}\hfill \, & \textbf{{17.46}\ms{0.38}}\hfill \,     \\
& CS-KD-E (ours)\hfill \,    & {{5.93}\ms{0.06}\hfill} \, & \textbf{{19.12}\ms{0.34}}\hfill \, & {{13.74}\ms{0.91}\hfill} \, & {{8.57}\ms{0.13}\hfill} \, & {{18.21}\ms{0.45}\hfill} \,     \\
\midrule
\multirow{7}{*}{ECE $\downarrow$} 
& Cross-entropy\hfill \,   &
{{15.45}\ms{0.33}\hfill} \, & {{14.08}\ms{0.76}\hfill} \, & {{18.39}\ms{0.76}\hfill} \, & {{15.05}\ms{0.35}\hfill} \, & {{17.99}\ms{0.72}\hfill} \,     \\
& AdaCos\hfill \,         & {{73.76}\ms{0.35}\hfill} \, & {{55.09}\ms{0.41}\hfill} \, & {{63.39}\ms{0.06}\hfill} \, & {{65.38}\ms{0.33}\hfill} \, & {{54.00}\ms{0.52}\hfill} \,     \\
& Virtual-softmax\hfill \, & {{8.02}\ms{0.55}\hfill} \, & {{4.60}\ms{0.67}\hfill} \, & {{11.68}\ms{0.66}\hfill} \, & {{7.91}\ms{0.38}\hfill} \, & {{11.21}\ms{1.00}\hfill} \,     \\
& Maximum-entropy\hfill \, & {{56.41}\ms{0.36}\hfill} \, & {{42.68}\ms{0.31}\hfill} \, & {{50.52}\ms{1.20}\hfill} \, & {{51.53}\ms{0.28}\hfill} \, & {{42.41}\ms{1.74}\hfill} \,     \\
& {Label-smoothing}\hfill \, & {{13.20}\ms{0.60}\hfill} \, & \textbf{{2.67}\ms{0.48}}\hfill \, & {{15.70}\ms{0.81}\hfill} \, & {{11.60}\ms{0.40}\hfill} \, & {{8.79}\ms{2.47}\hfill} \,     \\
% & \textcolor{red}{Clean Logit Pairing}\hfill \, & {{22.67}\ms{1.13}\hfill} \, & {{12.87}\ms{0.84}\hfill} \, & {{22.17}\ms{1.87}\hfill} \, & {{20.50}\ms{0.73}\hfill} \, & {{27.29}\ms{1.28}\hfill} \,     \\
& CS-KD (ours)\hfill \,    & {{5.17}\ms{0.40}\hfill} \, & {{7.26}\ms{0.93}\hfill} \, & {{15.44}\ms{0.92}\hfill} \, & {{10.46}\ms{1.08}\hfill} \, & {{15.56}\ms{0.29}\hfill} \,     \\
& CS-KD-E (ours)\hfill \,    & \textbf{{4.69}\ms{0.56}}\hfill \, & {{3.79}\ms{0.35}\hfill} \, & \textbf{{8.75}\ms{0.49}}\hfill \, & \textbf{{4.70}\ms{0.18}}\hfill \, & \textbf{{8.06}\ms{1.90}}\hfill \,     \\
\midrule
\multirow{7}{*}{R@1 $\uparrow$} 
& Cross-entropy\hfill \,   & {{61.38}\ms{0.64}\hfill} \, & {{30.59}\ms{0.42}\hfill} \, & {{33.92}\ms{1.70}\hfill} \, & {{47.51}\ms{1.02}\hfill} \, & {{31.42}\ms{1.00}\hfill} \,     \\
& AdaCos\hfill \,         & {{67.95}\ms{0.42}\hfill} \, & {{44.66}\ms{0.52}\hfill} \, & {{54.86}\ms{0.24}\hfill} \, & {{58.37}\ms{0.43}\hfill} \, & {{42.39}\ms{1.91}\hfill} \,     \\
& Virtual-softmax\hfill \, & {{68.35}\ms{0.48}\hfill} \, & {{44.69}\ms{0.58}\hfill} \, & {{55.56}\ms{0.74}\hfill} \, & {{59.71}\ms{0.56}\hfill} \, & {{44.20}\ms{0.90}\hfill} \,     \\
& Maximum-entropy\hfill \, & \textbf{{71.51}\ms{0.29}}\hfill \, & {{39.18}\ms{0.79}\hfill} \, & {{48.66}\ms{2.10}\hfill} \, & {{60.05}\ms{0.45}\hfill} \, & {{38.06}\ms{3.32}\hfill} \,     \\
& {Label-smoothing}\hfill \, & {{71.44}\ms{0.03}\hfill} \, & {{34.79}\ms{0.67}\hfill} \, & {{41.59}\ms{0.94}\hfill} \, & {{54.48}\ms{0.68}\hfill} \, & {{35.15}\ms{1.54}\hfill} \,     \\
% & \textcolor{red}{Clean Logit Pairing}\hfill \, & {{70.05}\ms{0.29}\hfill} \, & \textbf{{47.43}\ms{0.68}}\hfill \, & {{55.90}\ms{0.26}\hfill} \, & {{60.57}\ms{0.32}\hfill} \, & {{45.07}\ms{0.41}\hfill} \,     \\
& CS-KD (ours)\hfill \,    & {{71.15}\ms{0.15}\hfill} \, & \textbf{{47.15}\ms{0.40}\hfill} \, & \textbf{{59.06}\ms{0.38}}\hfill \, & \textbf{{62.67}\ms{0.07}}\hfill \, & \textbf{{46.74}\ms{1.48}}\hfill \,     \\
& CS-KD-E (ours)\hfill \,    & {{70.57}\ms{0.57}\hfill} \, & {{45.52}\ms{0.35}\hfill} \, & {{58.44}\ms{1.09}\hfill} \, & {{62.03}\ms{0.30}\hfill} \, & {{44.82}\ms{1.22}\hfill} \,     \\
\bottomrule
\end{tabular}}
\end{center}
\caption{Top-5 error, ECE, and Recall at 1 (R@1) rates (\%) of ResNet-18 on various image classification tasks. 
% We use ResNet-18 for fine-grained dataset, and PreAct ResNet-18 for CIFAR-100 and TinyImageNet datasets. 
We denote our method combined with the sample-wise regularization by CS-KD-E. 
The arrow on the right side of the evaluation metric indicates ascending or descending order of the value.
We reported the mean and standard deviation over three runs with different random seeds, and the best results are indicated in bold.}\label{tbl:ablation}
\vspace{-0.1in}
\end{table*}

\begin{figure*}[t]
\centering
\subfigure[\textcolor{black}{Cross-entropy}]
{\includegraphics[width=0.18\textwidth]{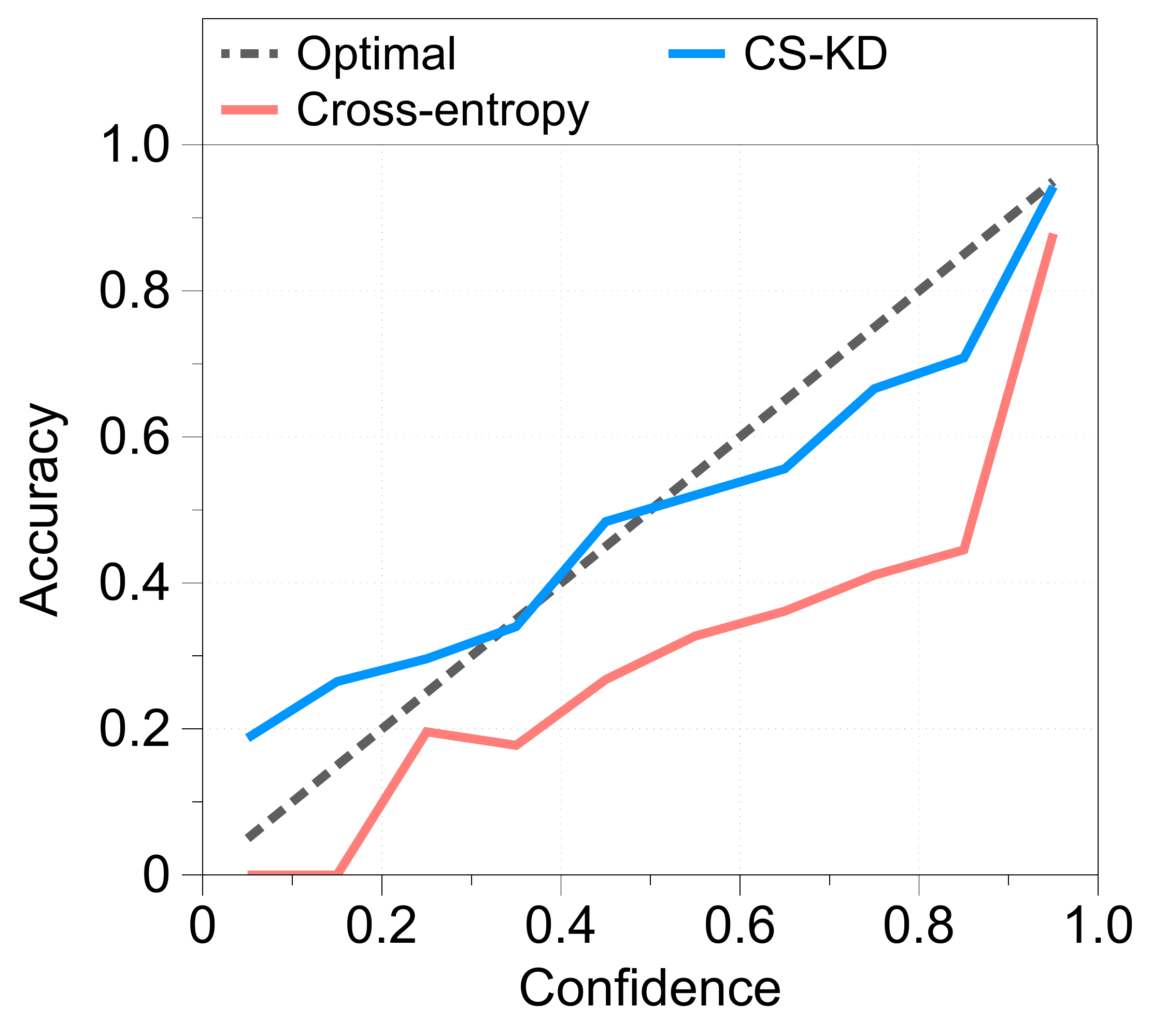}\label{fig:hist_ce}} 
\,
\subfigure[\textcolor{black}{Virtual-softmax}]
{
\includegraphics[width=0.18\textwidth]{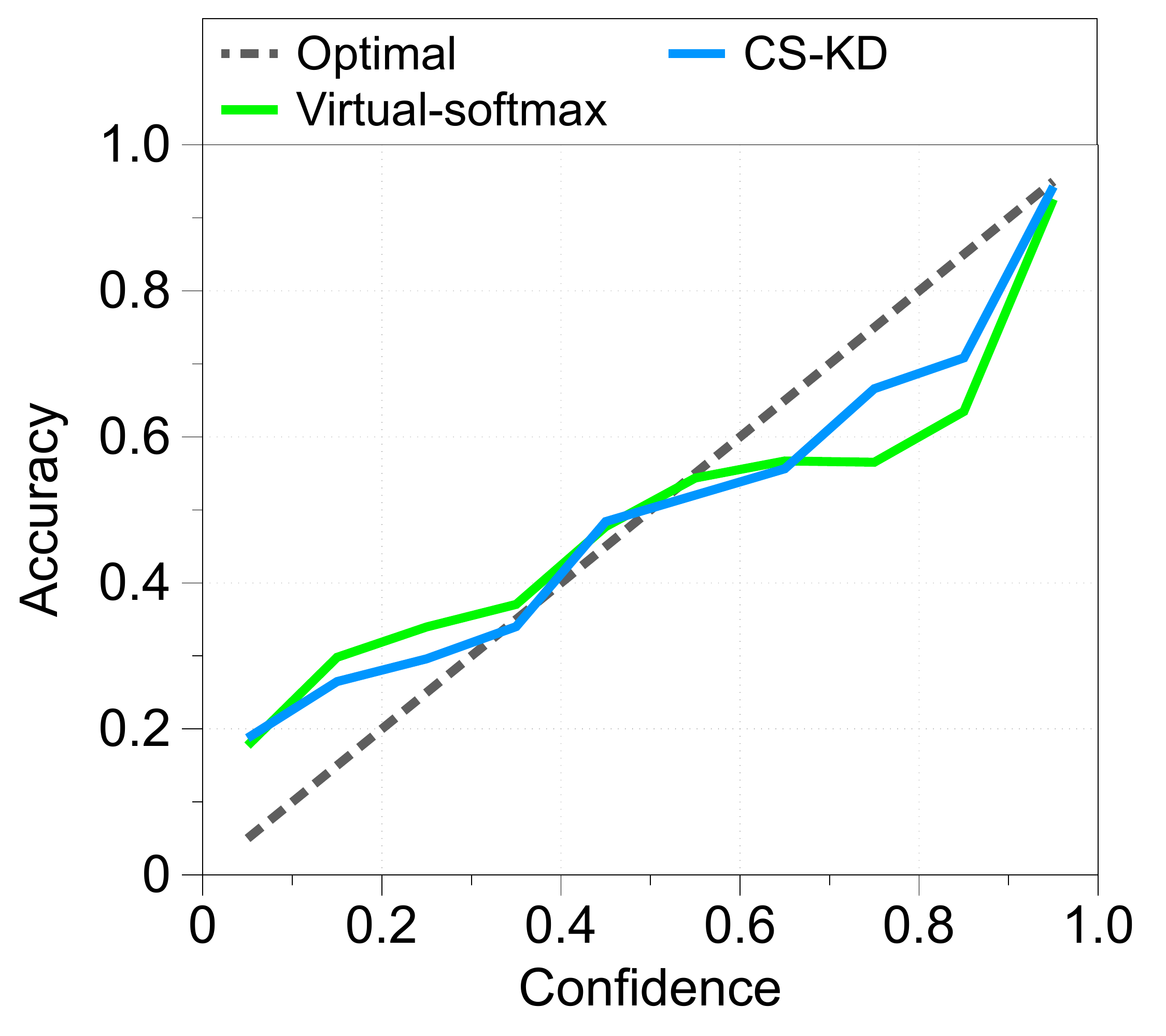}\label{fig:hist_vs}}
\,
\subfigure[\textcolor{black}{AdaCos}]
{
\includegraphics[width=0.18\textwidth]{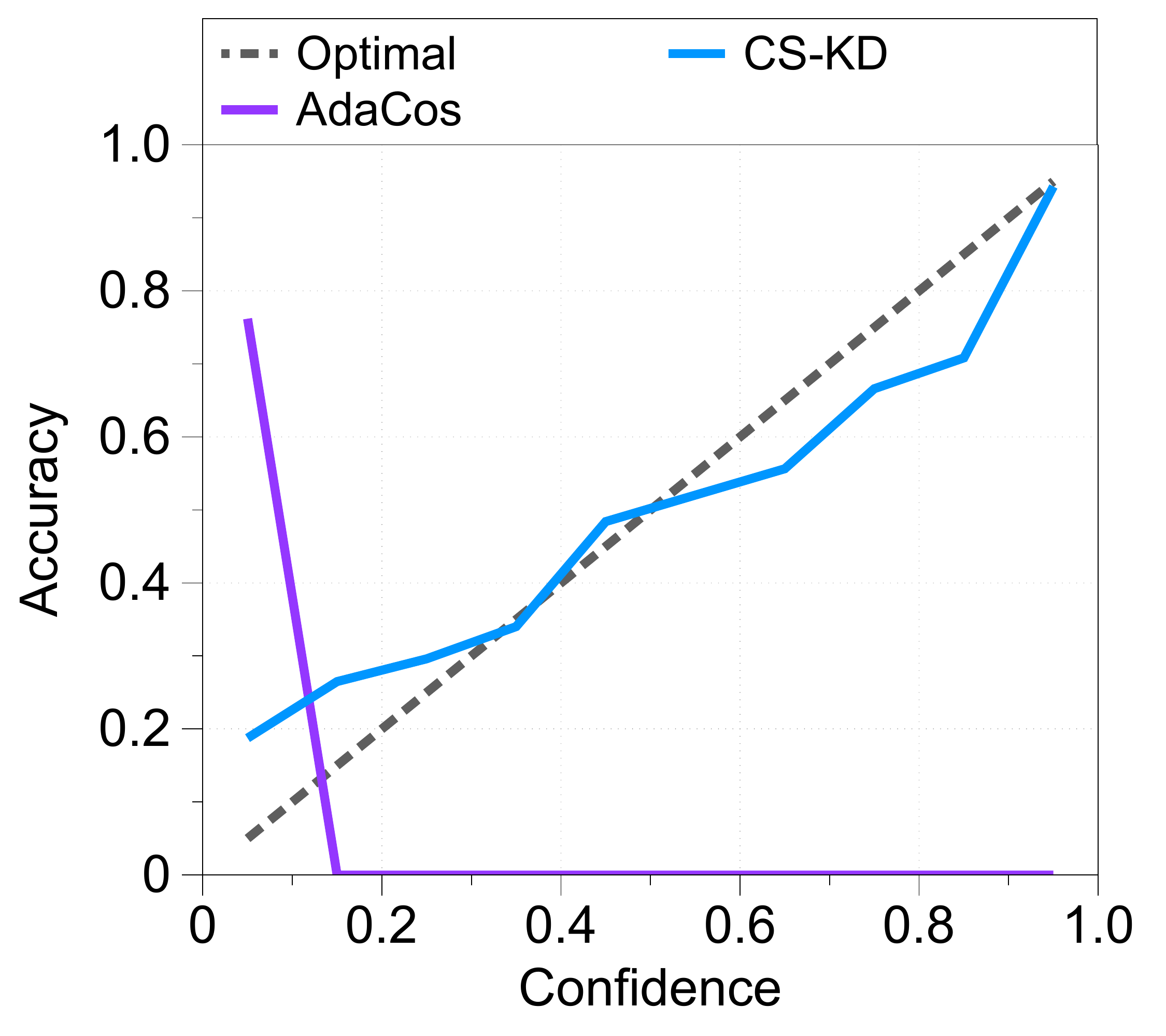}\label{fig:hist_adacos}}
\,
\subfigure[\textcolor{black}{Maximum-entropy}]
{
\includegraphics[width=0.18\textwidth]{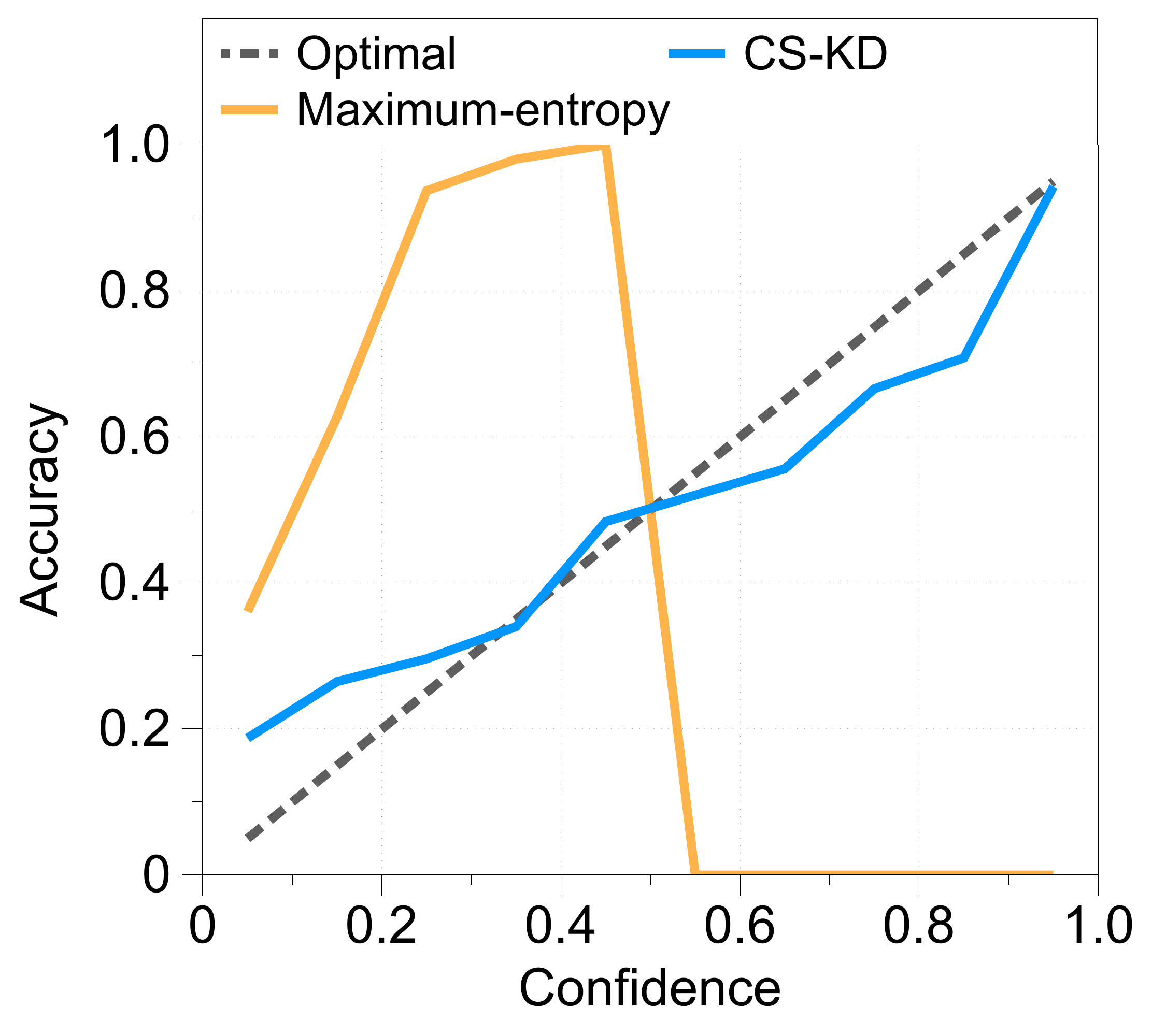}\label{fig:hist_me}} 
\subfigure[\textcolor{black}{Label-smoothing}]
{
\includegraphics[width=0.18\textwidth]{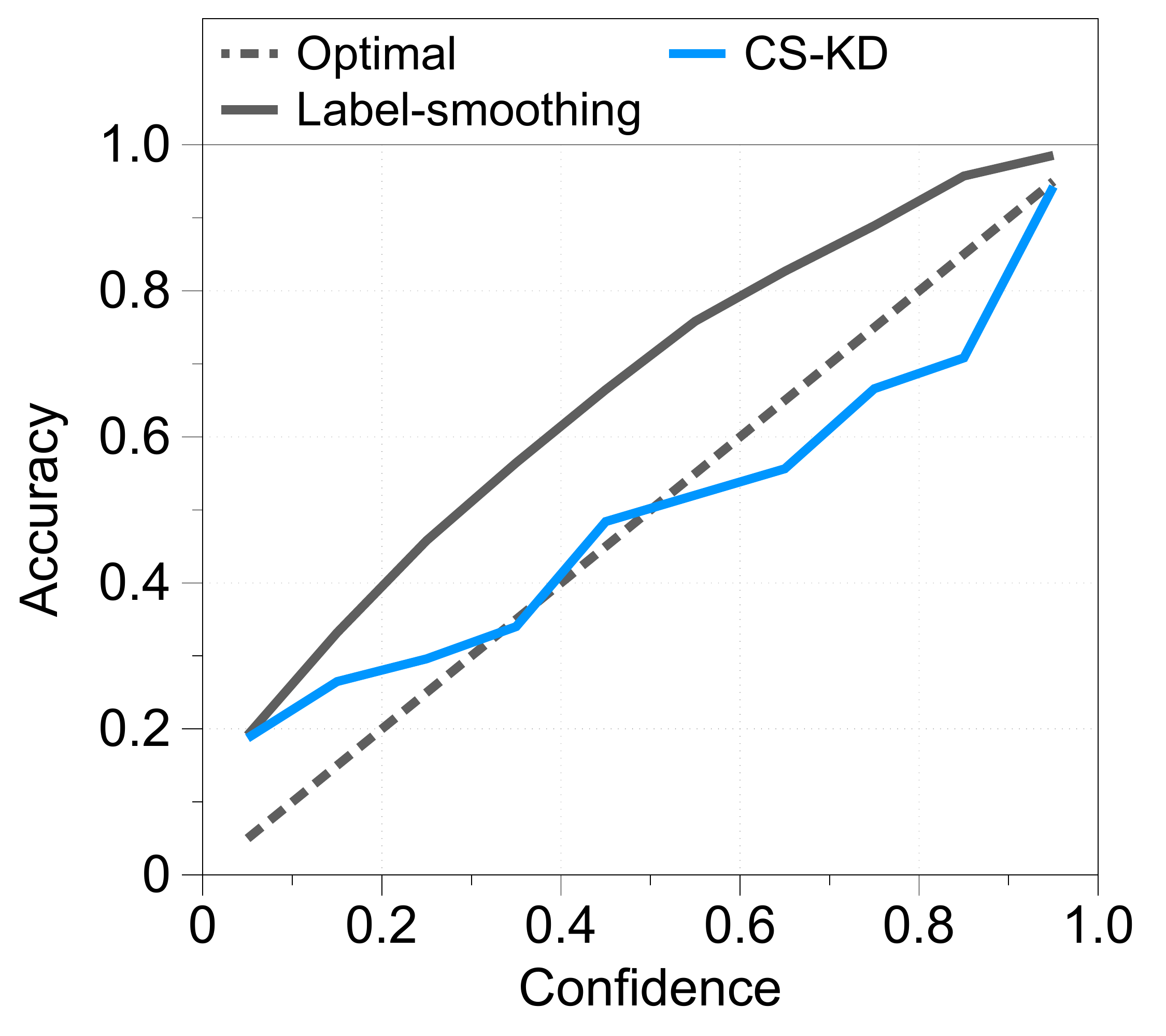}\label{fig:hist_ls}} 
% \,
% \subfigure[Mixup + CS-KD (ours)]
% {
% \includegraphics[width=0.3\textwidth]{figures/tSNE_c100_mxcskd.pdf} }
\caption{
\textcolor{black}{Reliability diagrams~\cite{degroot1983comparison, niculescu2005predicting} show accuracy as a function of confidence, for PreAct ResNet-18} trained on CIFAR-100 using (a) Cross-entropy, (b) Virtual-softmax, (c) AdaCos, (d) Maximum-entropy, and (e) Label-smoothing. All methods are compared with our proposed method, CS-KD. {Perfect calibration \cite{guo2017calibration} is plotted by dashed diagonals (Optimal) for all.}}
\label{fig:hist_all}
\vspace{-0.2in}
\end{figure*}

\subsection{Classification accuracy} \label{sec:classification}
%\vspace{-0.05in}
\noindent\textbf{Comparison with output regularization methods.}
We measure the top-1 error rates of the proposed method (denoted by CS-KD) by comparing with
Virtual-softmax, AdaCos, Maximum-entropy, and Label-smoothing on various image classification tasks. 
Table~\ref{tbl:main} shows that CS-KD outperforms other baselines consistently. 
In particular, 
CS-KD improves the top-1 error rate of the cross-entropy loss from 46.00\% to \textcolor{black}{33.28\%} under the CUB-200-2011 dataset.
We also observe that the top-1 error rates of other baselines are often worse than the cross-entropy loss, \eg, Virtual-softmax, Maximum-entropy, and {Label-smoothing} under
MIT67 and DenseNet).
%We also remark that top-5 error rates are also significantly improved due to meaningful predictions. 
As shown in Table~\ref{tbl:ablation}, top-5 error rates of \textcolor{black}{CS-KD} outperform other regularization methods, as it encourages meaningful predictions.
In particular, CS-KD improves the top-5 error rate of the cross-entropy loss from 6.91\% to {5.69\%} under the CIFAR-100 dataset, 
while the top-5 error rates of AdaCos is even worse than the cross-entropy loss.
%with 9.99\%.
These results imply that our method induces better predictive distributions than other baseline methods.

\vspace{0.05in}
\noindent\textbf{{Comparison with self-distillation methods.}}
%\textcolor{red}{
%Recently, self-distillation methods such as DDGSD \cite{xu2019data} and BYOT \cite{zhang2019your} are proposed. 
%Different from traditional knowledge distillation literature, the self distillation framework utilizes knowledge 
% within network 
%itself. 
%To verify the superiority of our methods, 
We also compare our method with recent proposed self-distillation techniques such as DDGSD \cite{xu2019data} and BYOT \cite{zhang2019your}.
As shown in Table~\ref{tbl:self},
CS-KD shows better top-1 error rates on ResNet-18 in overall.
%, while they have less effective results for some datasets. 
% In Table~\ref{tbl:self}, we compare our method with them on ResNet-18 using various datasets. 
% CS-KD shows improved top-1 error rates consistently for overall datasets, while they have less effective results for some datasets. 
For example, CS-KD shows the top-1 error rate of 33.28\% on the CUB-200-2011 dataset, while DDGSD and BYOT have 41.17\% and 40.76\%, respectively. 
%We remark that they are mainly benefited from the regularization effects of knowledge distillation. 
All tested self-distillation methods utilize regularization effects of knowledge distillation.
The superiority of CS-KD could be explained by its unique effect of reducing intra-class variations. %, which is a main component of .

\vspace{0.05in}
\noindent\textbf{{Evaluation on large-scale datasets.}}
{
To verify the scalability of our method, we have evaluated our method on the ImageNet
dataset with various model architecture such as ResNet-50, ResNet-101, and ResNeXt-101-32x4d \cite{xie2017aggregated}. 
As reported in Table \ref{tbl:imagenet}, our method improves 0.4\% of the
top-1 error rates across all the tested architectures consistently.
The 0.4\% improvement is comparable to, \eg, adding 51 more layers on ResNet-101 (\textit{i.e.}, ResNet-152) \cite{he2016deep}.}
% \textcolor{blue}{SM: one more sentence}
% \textcolor{blue}{Although not explored in this paper, we believe that our method is also compatible with any training scheme with advanced architectures.}

\vspace{0.05in}
\noindent\textbf{Compatibility with other regularization methods.}
We investigate orthogonal usage with other types of regularization methods
% compatibility of our method with other training methods 
such as Mixup \cite{zhang2017mixup} and knowledge distillation (KD) \cite{hinton2015distilling}.
%, for boosting the generalization. 
Mixup 
%is a regularization method that 
utilizes convex combinations of input pairs and corresponding label pairs for training.
We combine our method with Mixup regularization by applying 
the class-wise regularization loss $\mathcal{L}_{\tt{cls}}$ to mixed inputs and mixed labels, 
instead of standard inputs and labels. 
Table~\ref{tbl:mixup} shows the effectiveness of our method combined with Mixup regularization.
Interestingly, this simple idea significantly improves the performances of fine-grained classification tasks.
%Table~\ref{tbl:mixup} shows that  Experiments are in Table~\ref{tbl:mixup} and and follow other experimental setups in Section~\ref{exp:setup}. 
% \textcolor{blue}{
In particular, 
our method improves the top-1 error rate of Mixup regularization from 37.09\% to \textcolor{black}{30.71\%}, where the top-1 error rate of the cross-entropy loss is 46.00\% under ResNet-18 on the CUB-200-2011 dataset.
% our method with Mixup regularization achieves 14.05\% of absolute improvements compared to cross-entropy loss, where Mixup and our regularization achieve 5.91\% and 12.5\% of absolute improvements in CUB-200-2011 dataset, respectively.}

KD regularizes predictive distributions of student network to learn the dark knowledge of a teacher network. 
% We combine our method with KD by applying a teacher network to regularize predictive distributions learning knowledge from teacher and itself simultaneously.
We combine our method with KD to learn dark knowledge from the teacher and itself simultaneously.
%Early stage of training, teacher knowledge dominates self-knowledge, because self-knowledge is poorly generalized at the beginning. However, self-knowledge eventually outperforms teacher knowledge, because teacher knowledge is fixed. 
Table~\ref{tbl:kd} shows that our method achieves a similar performance of KD, although ours does not use additional teacher networks. 
Besides, combining our method with KD further improves the top-1 error rate of our method from 39.32\% to \textcolor{black}{34.47\%}, where the top-1 error rate of the cross-entropy loss is 48.36\% under ResNet-10 trained on the CUB-200-2011 dataset.
%We expect other types of regularization methods using the teacher network also effective to improve our method.
These results show the wide applicability of our method, compatible to use with other regularization methods.

% \vspace{-0.05in}
\subsection{Ablation study}\label{sec:analysis}
%\vspace{-0.05in}
\noindent\textbf{Feature embedding analysis.} 
One can expect that the intra-class variations can be reduced by forcing DNNs to produce meaningful predictions.
To verify this, we analyze the feature embedding of the penultimate layer of ResNet-18 trained on CIFAR-100 dataset by t-SNE~\cite{maaten2008visualizing} visualization method.
As shown in Figure \ref{fig:tsne_all}, the intra-class variations are significantly decreased by our method (Figure~\ref{fig:tsne_cskd}) compared to other baselines, including Virtual-softmax (Figure~\ref{fig:tsne_vs}) and AdaCos (Figure~\ref{fig:tsne_adacos}), which are designed to reduce intra-class variations.
%, while only reduce the angular-margin. 
We also provide quantitative results on the metric Recall at 1 (R@1), which has appeared in Section~\ref{exp:setup}. We remark that the larger value of R@1 implies small intra-class variations on the feature embedding \cite{wengang2017recent}.
% various metrics such as R@1 to analyze feature embedding in Table~\ref{tbl:ablation}. 
As shown in Table~\ref{tbl:ablation},
R@1 values can be significantly improved when ResNet-18 is trained by our method.
In particular, R@1 of \textcolor{black}{CS-KD} is 47.15\% under the TinyImageNet dataset, while R@1 of 
Adacos, Virtual-softamx, and
the cross-entropy loss are 44.66\%, 44.69\%, and 30.59\%, respectively.

\vspace{0.05in}
\noindent\textbf{Hierarchical image classification.}
By producing more semantic predictions, \textit{i.e.}, increasing the correlation between similar classes in predictions,
we expect the trained classifier can capture a hierarchical (or clustering) structure of label space.
To verify this, we evaluate the proposed method on a mixed dataset with 387 fine-grained labels and three hierarchy labels, \textit{i.e.}, bird (CUB-200-2011; 200 labels), dog (Stanford Dogs; 120 labels), and indoor (MIT67; 67 labels).
Specifically, 
we randomly choose 30 samples per each fine-grained label for training, and original test datasets are used for the test.
For evaluation, we train ResNet-18 to classify the fine-grained labels and measure a hierarchical classification accuracy by predicting a hierarchy label (bird, dog, or indoor) as that of predicted fine-grained label.

First, we extract the hierarchical structure as confusion matrices, where each element indicates the hierarchical image classification accuracy.
As shown in Figure~\ref{fig:matrix:ce} and \ref{fig:matrix:kl},
our method captures the hierarchical structure of the mixed dataset almost perfectly, \textit{i.e.}, showing the identity confusion matrix. In particular, our method enhances the hierarchical image classification accuracy significantly up to 99.3\% in the bird hierarchy (CUB-200-2011). 
Moreover, as shown in Figure~\ref{tbl:conf}, our method also improves the top-1 error rates of fine-grained label classification significantly. Interestingly, the error rate of CUB-200-2011 is even lower than the errors reported in Table~\ref{tbl:main}. This is because the model learns additional information by utilizing the dark knowledge of more labels.

\subsection{Calibration effects} \label{sec:calibration}
%\vspace{-0.05in}
%In this section, we compare the methods with various performance metrics for showing that the proposed regularization method indeed encourages learning better predictive distributions.
In this section, we also evaluate the calibration effects of the proposed regularization method.
%One can expect the model producing meaningful predictions is well-calibrated.
Specifically, 
we provide reliability diagrams~\cite{degroot1983comparison, niculescu2005predicting}, which plot the expected sample accuracy as a function of confidence of PreAct ResNet-18 for the CIFAR-100 dataset in Figure~\ref{fig:hist_all}.
We remark that the plotted identity function (dashed diagonal) implies perfect calibration \cite{guo2017calibration}, and our method is the closest one among the baselines, as shown in Figure \ref{fig:hist_all}. 
Moreover, we evaluate our method by ECE~\cite{guo2017calibration,naeini2015obtaining}, which is a quantitative metric of calibration, in Table~\ref{tbl:ablation}. The results demonstrate that our method outperforms the cross-entropy loss consistently. In particular, CS-KD enhances ECE of the cross-entropy from {15.45\% to 5.17\% under the CIFAR-100 dataset, while AdaCos and Maximum-entropy are significantly worse than the cross-entropy with 73.76\% and 56.41\%}, respectively.

% One can 
%straightforward extension of the proposed method, 
%\textit{i.e.}, forcing outputs on a given sample and its perturbed one should be similar.
{As a natural extension of CS-KD, we also consider combining our method with an existing consistency loss \cite{bachman2014learning, clark2018semi, miyato2018virtual, xie2019uda,tarvainen2017mean},
which regularizes the output distributions of a given sample and its augmented one.}
% We remind that consistency regularizers have been widely applied in the literature .
% Interestingly, we found that our method combined with the sample-wise regularization (CS-KD-E) enhances the model calibration.} %learning semantically better predictive distributions. 
%Remind that this kind of techniques has been employed widely in the semi-supervised learning.
% To employ the sample-wise regularization, we adopted an additional loss term similar to the unsupervised data augmentation \cite{xie2019uda} to $\mathcal{L}_{\tt{\tt CS{\text -}KD}}$.
Specifically, for a given training sample $\mathbf{x}$ and another sample $\mathbf{x}^\prime$ having the same label, the combined regularization loss $\mathcal{L}_{\tt{\tt CS{\text -}KD{\text -}E}}$ is defined as follows:
%In addition to enforcing the intra-class consistency of predictive distributions, we apply this idea to the single-sample scenario by augmenting the input data.
%\vspace{-0.05in}
\begin{align*}
    \mathcal{L_{\tt{\tt CS{\text -}KD{\text -}E}}}(\mathbf{x}, \mathbf{x}^\prime, y; \theta, T) & :=
      \ \mathcal{L_{\tt{\tt CS{\text -}KD}}}(\mathbf{x}_{\tt{aug}}, \mathbf{x}_{\tt{aug}}^\prime, y; \theta, T)\notag \\
      + \ \lambda_{\tt E} \cdot T^2 \cdot &\text{KL}\left(P(y|\mathbf{x};\widetilde{\theta},T)
    \big\| 
    P(y|\mathbf{x}_{\tt{aug}};{\theta},T)\right),  \label{lam:cls} 
\end{align*}
where $\mathbf{x}_{\tt{aug}}$ is an augmented sample that is generated by 
the data augmentation technique\footnote{We use standard data augmentation techniques (\textit{i.e.}, flipping and random sized cropping) for all tested methods in this paper.}, and $\lambda_{\tt E} >0$ is the loss weight for balancing.
The corresponding results are reported in
Table~\ref{tbl:ablation}, denoted by CS-KD-E.
% For fine-grained datasets such as CUB-200-2011, Stanford Dogs and MIT67, CS-KD-E improves not only classification accuracy, but also enhances the calibration and R@1 significantly. 
%reliability diagrams in Figure~\ref{fig:hist_all}
We found that CS-KD-E significantly enhances the calibration performance of CS-KD, and also outperforms the baseline methods over top-1 and top-5 error rates consistently. 
% with only slight increase of error rates than CS-KD. 
% We note that CS-KD-E achieves still much better error rates and R@1 than the cross-entropy. 
%This implies that sample-wise regularization could be also useful when the number of samples per class is small.
In particular, CS-KD-E enhances ECE of CS-KD from {5.17\% to 4.69\% under the CIFAR-100} dataset.
We think that investigating the effect of such combined regularization could be an interesting direction to explore in the future, \eg, utilizing other augmentation methods such as 
% resizing, rotating, random cropping~\cite{krizhevsky2009learning, simonyan2014very, szegedy2015going}, 
cutout~\cite{devries2017improved} and auto-augmentation~\cite{cubuk2018autoaugment}.

%\vspace{-0.05in}
\section{Related work}
%\vspace{-0.05in}
\noindent\textbf{Regularization techniques.} 
Numerous techniques have been introduced to prevent overfitting of neural networks, including early stopping \cite{bishop1995regularization}, $L_1$/$L_2$-regularization \cite{nowlan1992simplifying}, dropout \cite{srivastava2014dropout}, and batch normalization \cite{ioffe2015batch}. 
%These techniques control the parameters of a neural network. 
Alternatively, regularization methods for the predictive distribution also have been explored: 
% Numerous regularization methods for the predictive distribution also have been explored: 
Szegedy \etal \cite{szegedy2016rethinking} proposed label-smoothing, which is a mixture of the ground-truth and the uniform distribution, and Zhang \etal \cite{zhang2017mixup} proposed a data augmentation method called Mixup, which linearly interpolates a random pair of training samples and corresponding labels. 
{M{\"u}ller \etal \cite{muller2019does} investigated a method called Label-smoothing and empirically showed that it improves not only generalization but also model calibration in various tasks, such as image classification and machine translation.}
Similarly, Pereyra \etal \cite{pereyra2017regularizing} proposed penalizing low entropy predictive distribution, which improved exploration in reinforcement learning and supervised learning. 
Moreover, several works \cite{bachman2014learning, clark2018semi, xie2019uda,tarvainen2017mean} investigated consistency regularizers between the predictive distributions of corrupted samples and original samples for semi-supervised learning. 
%{Logit pairing \cite{kannan2018adversarial} is a method that encourages logits for pairs of samples to be similar. It was originally proposed for adversarial robustness, which improves accuracy on adversarial examples over adversarial training.
%by minimizing distance between logits for clean examples and their adversarial counterparts.
%In a clean version of logit pairing, the pairs are two randomly selected clean training examples. 
%}
We remark that our method enjoys orthogonal usages with the prior methods, \textit{i.e.}, our method can be combined with the prior methods to further improve the generalization performance.

\vspace{0.05in}
\noindent\textbf{Knowledge distillation.}
Knowledge distillation \cite{hinton2015distilling} is an effective learning method to transfer the knowledge from a powerful teacher model to a student. 
This pioneering work showed that one can use softmax with temperature scaling to match soft targets for transferring {\em dark knowledge}, which contains the information of non-target labels. %Therefore knowledge distillation often used to compress the information to a relatively small neural network.
There are numerous follow-up studies to distill knowledge in the aforementioned teacher-student framework. 
% FitNets \cite{romero2014fitnets} tried to learn features of a thin deep network using a shallow one with linear transform. Similarly, \cite{zagoruyko2016paying} introduced a transfer method that matches attention maps of the intermediate features, and \cite{ahn2019variational} tried to maximize the mutual information between intermediate layers of teacher and student for enhanced performance. \cite{srinivas2018knowledge} proposed a loss function for matching Jacobian of the network’s output instead of the feature itself.
{Recently, 
some of the self-distillation approaches \cite{xu2019data, zhang2019your}, 
% there are some recent self-distillation approaches 
which distill knowledge itself,
are proposed. 
Data-distortion guided self-distillation method \cite{xu2019data} transfers knowledge between different augmented versions of the same training data.
%similar to consistency regularization methods. 
Be Your Own Teacher \cite{zhang2019your}, on the other hand, utilizes ensembling predictions from multiple branches to improve its performance.}
We remark that our method and 
{these knowledge distillation methods 
have a similar component, \textit{i.e.}, using a soft target distribution, but ours 
only reduces intra-class variations.}
% utilizes the soft target distribution from itself. 
We also remark that the joint usage of our method and the prior knowledge distillation methods is also possible.
% self distillation
%Recently, (Hahn et al.) proposed a training method using second soft prediction of the model itself for natural langulage processing (NLP), called self-knowledge distillation. Compared to (Hahn et al.), this paper heads from a different idea that encourages intra-class consistency of predictions via knowledge distillation using full predictions, without loss of information.

\vspace{0.05in}
\noindent\textbf{Margin-based softmax losses.}
There have been recent efforts toward boosting the recognition performances via enlarging inter-class margins and reducing intra-class variation. Several approaches utilized metric-based methods that measure similarities between features using Euclidean distances, such as triplet \cite{weinberger2009distance} and contrastive loss \cite{chopra2005learning}. To make the model extract discriminative features, center loss \cite{wen2016discriminative} and range loss \cite{zhang2017range} were proposed to minimize distances between samples belong to the same class. 
% COCO loss \cite{liu2017learning} and NormFace \cite{wang2017normface} optimized cosine similarities, by utilizing reformulations of softmax loss and metric learning with feature normalization. Similarly, \cite{zheng2018ring} applied ring loss for soft normalization which uses a convex norm constraint.
%It is widely applied in computer vision tasks, such as face verification \cite{schroff2015facenet} and identification \cite{cheng2016person}, and fine-grained classification \cite{wang2016mining}.
Recently, angular-margin based losses were proposed for further improvement. L-softmax \cite{liu2016large} and 
A-softmax \cite{liu2017sphereface} combined angular-margin constraints with softmax loss to encourage the model to generate more discriminative features. CosFace \cite{wang2018cosface}, AM-softmax \cite{wang2018amsoft}, and ArcFace \cite{deng2019arcface} introduced angular-margins for a similar purpose, by reformulating softmax loss. Different from L-Softmax and A-Softmax, Virtual-softmax \cite{chen2018virtual} encourages a large margin among classes via injecting additional virtual negative class.

%\vspace{-0.05in}
\section{Conclusion}
%\vspace{-0.05in}
In this paper, we discover a simple regularization method to enhance the generalization performance of deep neural networks. 
We propose the regularization term, which penalizes the predictive distribution between different samples of the same label by minimizing the Kullback-Leibler divergence.
We remark that our idea regularizes the dark knowledge (\textit{i.e.}, the knowledge on wrong predictions) itself and encourages the model to produce more meaningful predictions.
Moreover, we demonstrate that our proposed method can be useful for the generalization and calibration of neural networks.
We think that the proposed regularization technique would enjoy a broader range of 
applications, such as exploration in deep reinforcement learning \cite{haarnoja2018soft}, transfer learning \cite{ahn2019variational}, face verification \cite{deng2019arcface}, and detection of out-of-distribution samples \cite{lee2018training}. 

%\vspace{-0.05in}
\section*{Acknowledgments}
%\vspace{-0.05in}
{
This work was supported by Institute for Information \& communications Technology Promotion (IITP) grant funded by the Korea government (MSIT) 
(No.2016-0-00563, Research on Adaptive Machine Learning Technology Development for Intelligent Autonomous Digital Companion) 
% and Korea Evaluation Institute of Industrial Technology(KEIT) grant funded by the Korea government(MOTIE), 
% This work was partly supported by 
and 
Institute of Information \& communications Technology Planning \& Evaluation (IITP) grant funded by the Korea government (MSIT) 
(No.2019-0-00075, Artificial Intelligence Graduate School Program (KAIST)). 
% and Korea Evaluation Institute of Industrial Technology(KEIT) grant funded by the Korea government(MOTIE). 
We also
thank Sungsoo Ahn and Hankook Lee 
for helpful discussions.}

{\small
\bibliographystyle{ieee_fullname}
\bibliography{egbib}
}

\newpage
\appendix

\onecolumn
    \clearpage
    \begin{center}{\bf {\LARGE Supplementary Material:}}
    \end{center}
    \begin{center}{\bf {\Large Regularizing Class-wise Predictions via Self-knowledge Distillation}}
    \end{center}

\vspace{0.1in}

\section{{Effects of hyper-parameters}}% Effect of hyperparamters $T$ and $\lambda_{\tt cls}$.}} 
{
To examine the effect of main hyper-parameters $T$ and $\lambda_{\tt cls}$, 
we additionally test the hyper-parameters
across an array of $T\in\{0.1, 0.5, 1, 4, 10, 20\}$ and $\lambda_{\tt cls}\in\{0.1, 0.5, 1, 2, 3, 4, 10, 20\}$ on PreAct ResNet-18 using the CIFAR-100 dataset.
The results are presented in Table \ref{tbl:hyper}. Except for the hyper-parameters under consideration, we keep all settings the same as in Section 3.1. %\ref{exp:setup}. 
Overall, we found our method 
is fairly robust on $T$ and $\lambda_{\tt cls}$, except for some extreme cases, such as the small value of $T\leq0.5$, and the large value of $\lambda_{\tt cls} \geq 10$.}

\begin{table*}[h]
\begin{center}
\begin{tabular}{c|cccccccc} 
\toprule
\diagbox{$T$}{$\lambda_{\tt cls}$} & { 0.1}& {0.5} & {1} & {2} & {3} & {4} & {10} & {20} \\ 
\midrule
0.1 &  25.16 & 24.03 & 23.91 & 24.38 & 24.05 & 24.21 & 24.39 & 27.61\\
0.5 &  24.14 & 24.05 & 24.15 & 23.49 & 23.78 & 23.23 & 23.90 & 25.96\\
{1}   &  24.15 & 23.32 & 22.80 & 22.26 & 22.87 & 23.18 & 24.35 & 25.58\\
4   &  22.87 & 22.03 & \textbf{21.66} & 22.45 & 22.68 & 22.81 & 32.25 & 35.45\\
10  &  22.68 & 22.36 & {21.98} & 22.04 & 21.95 & 31.76 & 31.80 & 37.50\\
20  &  22.96 & 22.39 & 22.03 & 22.37 & 22.00 & 22.39 & 30.23 & 24.05\\
\bottomrule
\end{tabular}
\end{center}
\vspace{0.1in}
\caption{{Top-1 error rates (\%) of PreAct ResNet-18 on CIFAR-100 dataset over various hyper-parameters $T$ and $\lambda_{\tt cls}$. 
The best results are indicated in bold.}}\label{tbl:hyper}
\end{table*}

\section{{Qualitative analysis of CS-KD}}
\begin{figure*}[h]
\centering
%\subfigure[{Log-probabilities of predicted labels on misclassified samples}]
{
\includegraphics[width=0.48\textwidth]{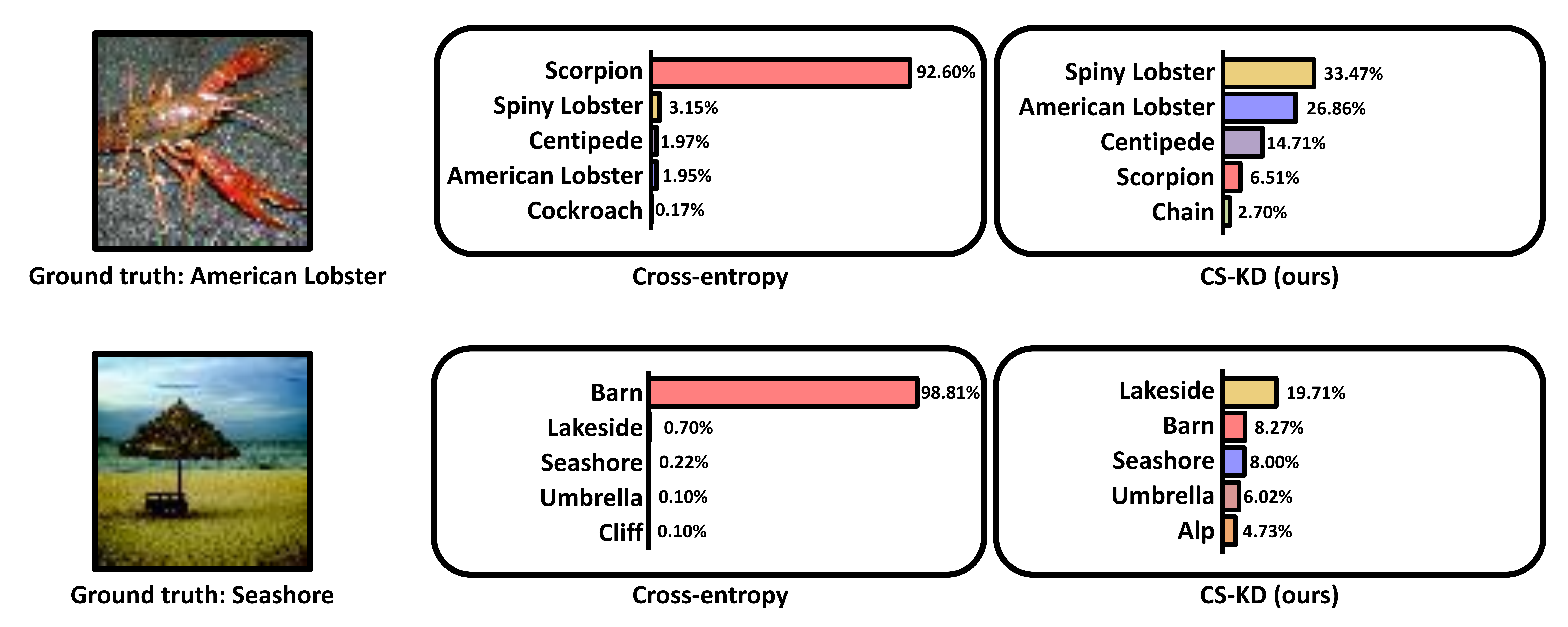}} 
\,
%\subfigure[{Log-probabilities of ground-truth labels on misclassified samples}]
{
\includegraphics[width=0.48\textwidth]{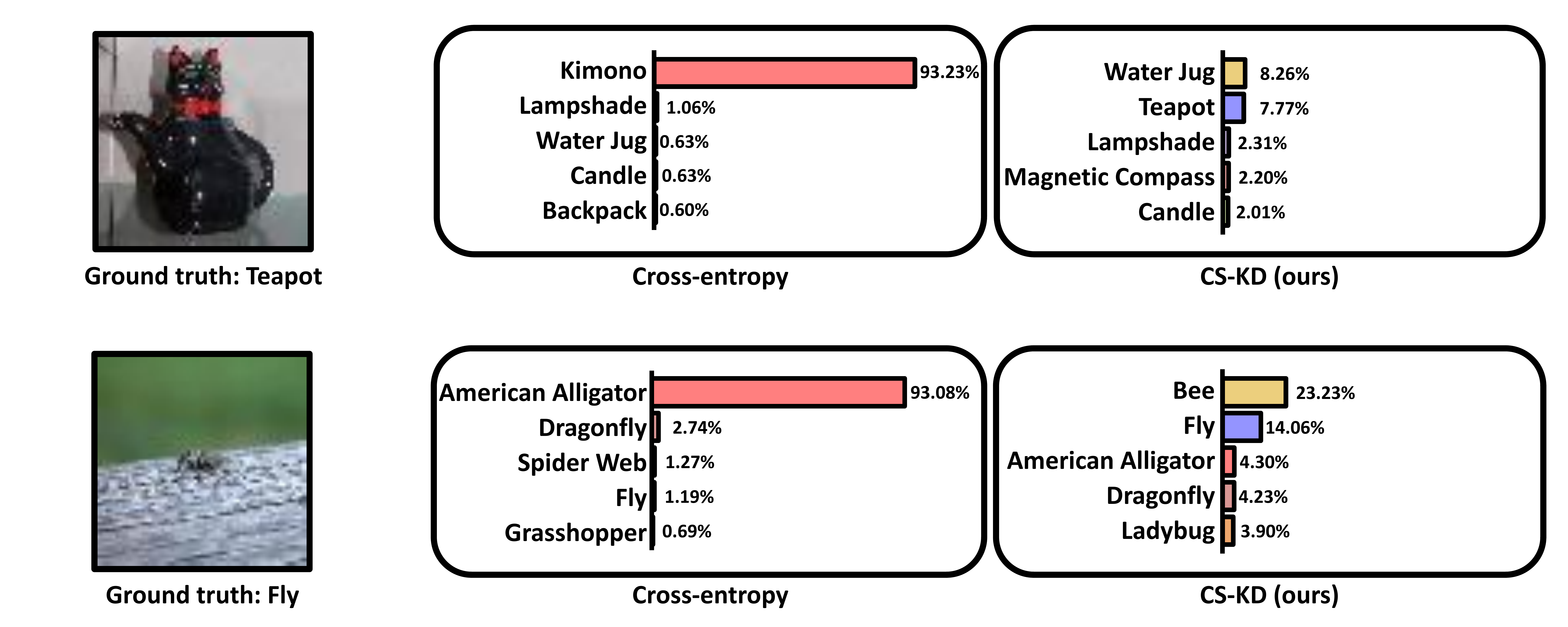}}
%\vspace{0.1in}
\caption{Predictive distributions on misclassified samples. We use PreAct ResNet-18 trained on TinyImageNet dataset. 
For misclassified samples, softmax scores of the ground-truth class are increased by training DNNs with class-wise regularization.}\label{fig:samples_tiny}
\end{figure*}

To examine the effect of our method, 
we investigate prediction values in softmax scores, \textit{i.e.}, $P(y|\mathbf{x})$, 
from PreAct ResNet-18 trained by the standard cross-entropy loss and our method for TinyImageNet dataset.
We report commonly misclassified samples by both the cross-entropy and our method in Figure~\ref{fig:samples_tiny}, 
% As shown in Figure~\ref{fig:samples_tiny}, samples are commonly misclassified by both the cross-entropy and our method, 
and softmax scores of the samples show our method not only moderates the overconfident predictions, but also enhances the prediction values of classes correlated to the ground-truth class.

\begin{figure*}[t]
\centering
\vspace{-0.30in}
\subfigure[{Log-probabilities of predicted labels on misclassified samples}]
{
\includegraphics[width=0.46\textwidth]{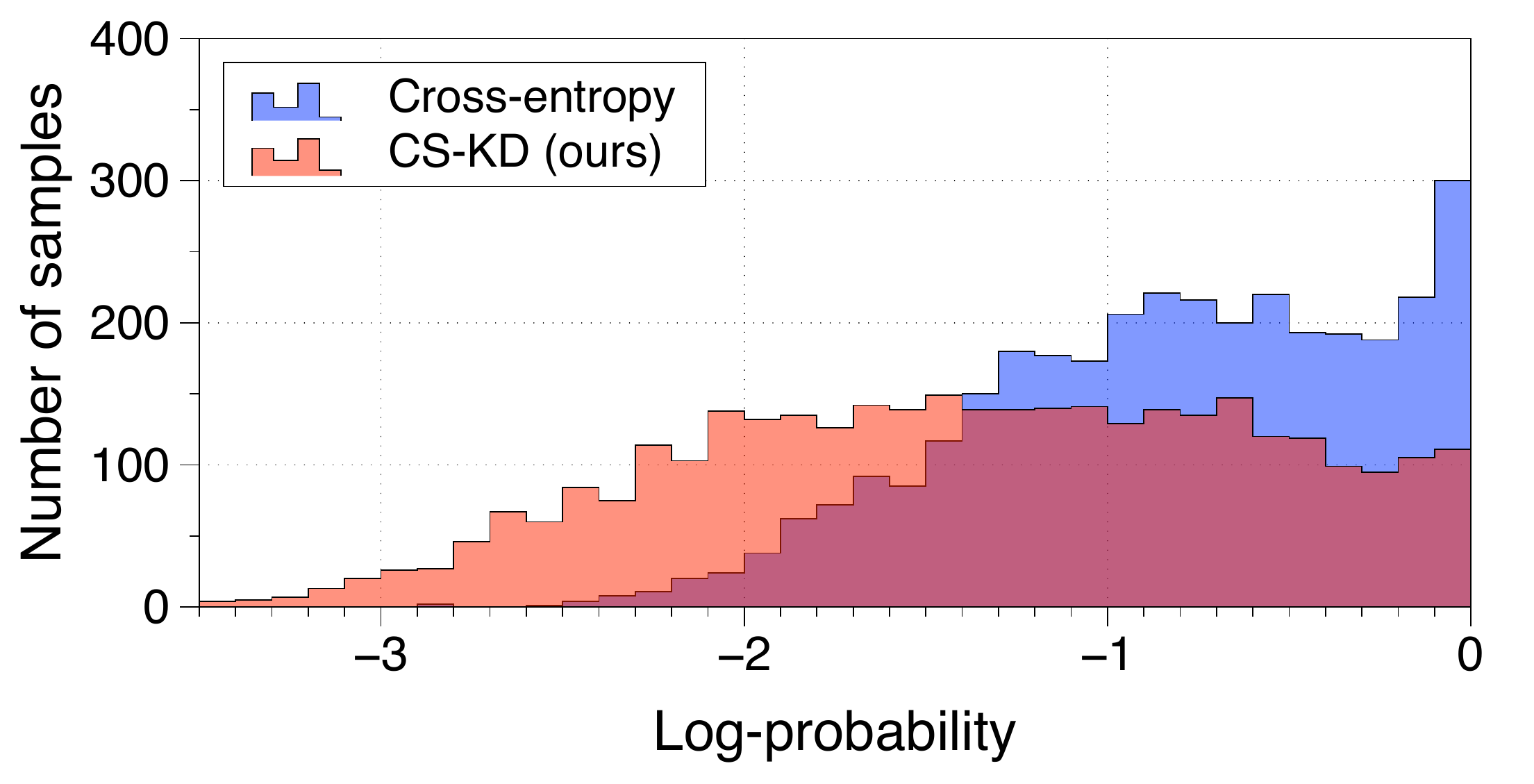}} 
\,
\subfigure[{Log-probabilities of ground-truth labels on misclassified samples}]
{
\includegraphics[width=0.46\textwidth]{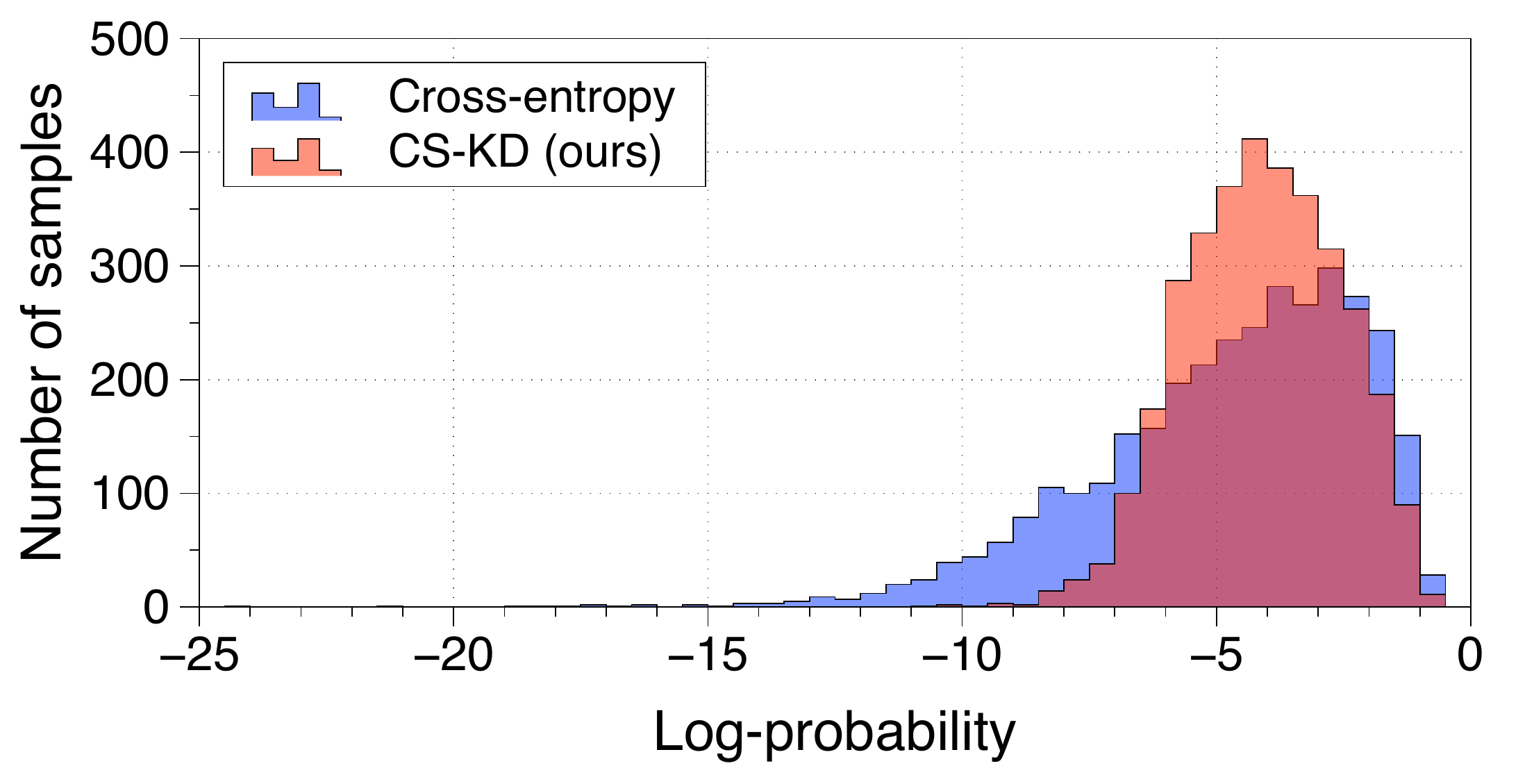}}
\caption{\textcolor{black}{Histogram of log-probabilities of (a) the predicted label, \textit{i.e.}, top-1 softmax score, and (b) the ground-truth label on misclassified samples by networks trained by {the} cross-entropy {(baseline)} and CS-KD.
The networks are trained on PreAct ResNet-18 for TinyImageNet.}}\label{hist:tiny}

\vspace{-0.05in}
\subfigure[\textcolor{black}{Log-probabilities of predicted labels on misclassified samples}]
{
\includegraphics[width=0.46\textwidth]{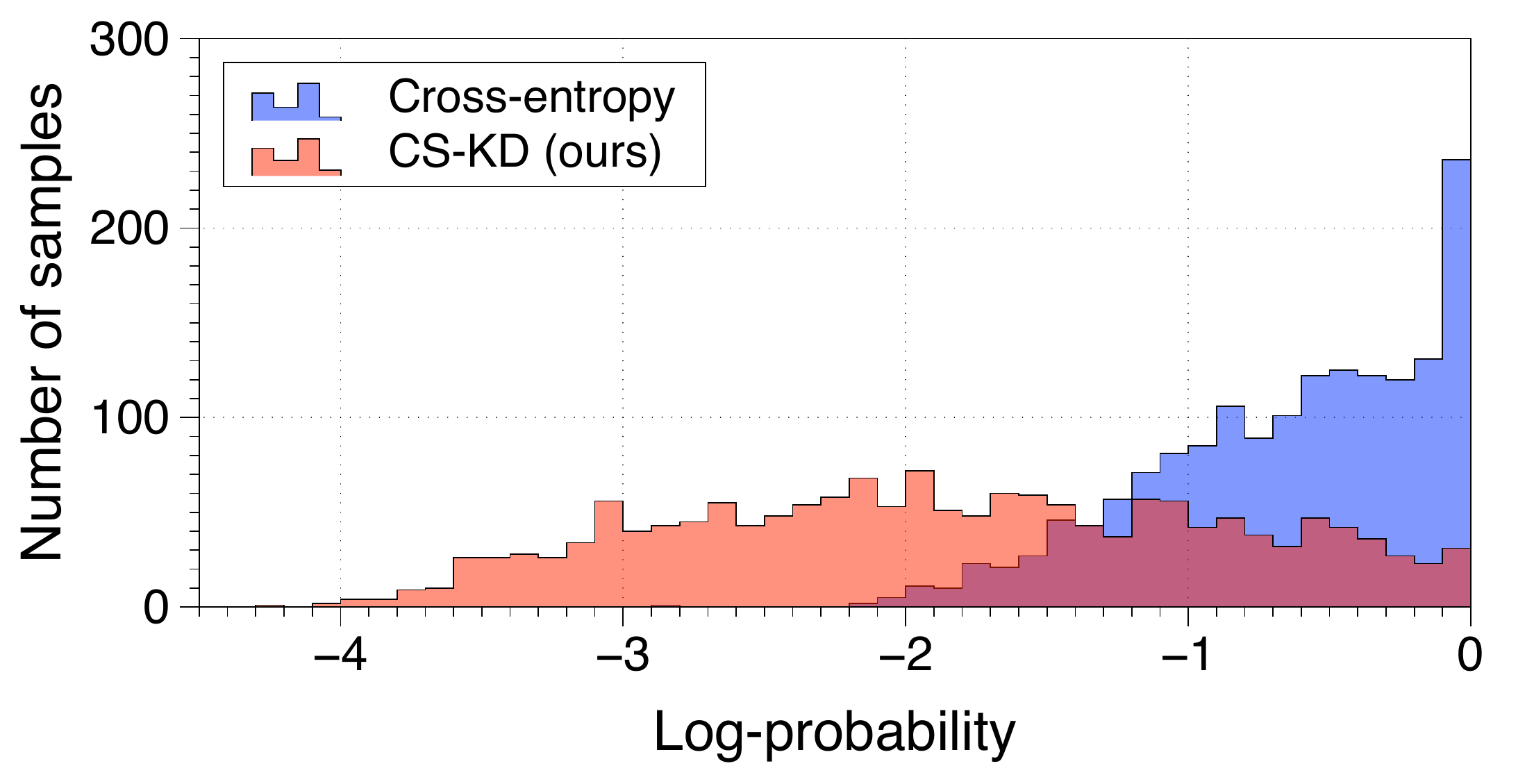}} 
\,
\subfigure[\textcolor{black}{Log-probabilities of ground-truth labels on misclassified samples}]
{
\includegraphics[width=0.46\textwidth]{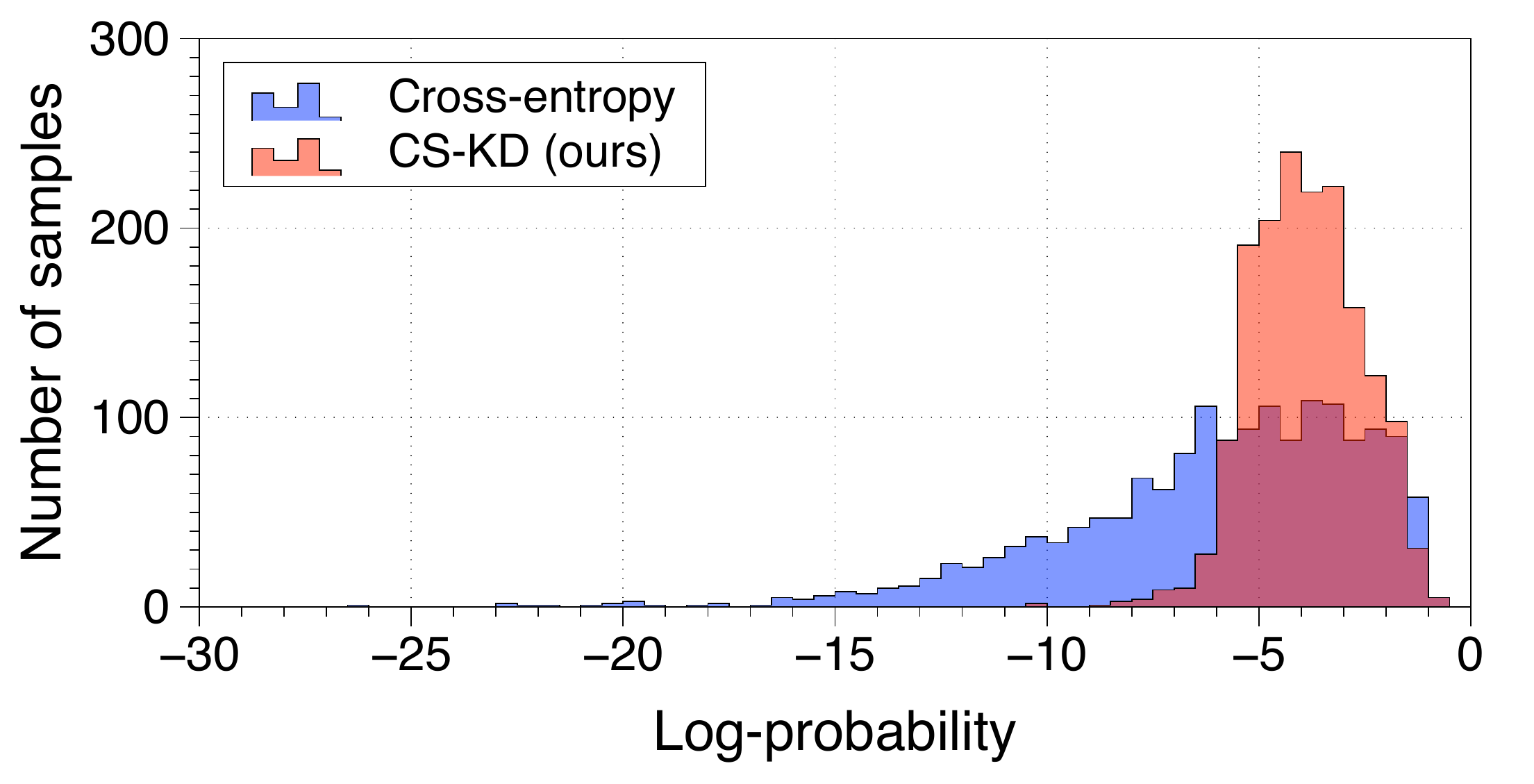}}
\caption{\textcolor{black}{Histogram of log-probabilities of (a) the predicted label, \textit{i.e.}, top-1 softmax score, and (b) the ground-truth label on misclassified samples by networks trained by {the} cross-entropy {(baseline)} and CS-KD.
The networks are trained on ResNet-18 for CUB-200-2011.}}\label{hist:cub}

% \vspace{-0.05in}
\subfigure[\textcolor{black}{Log-probabilities of predicted labels on misclassified samples}]
{
\includegraphics[width=0.46\textwidth]{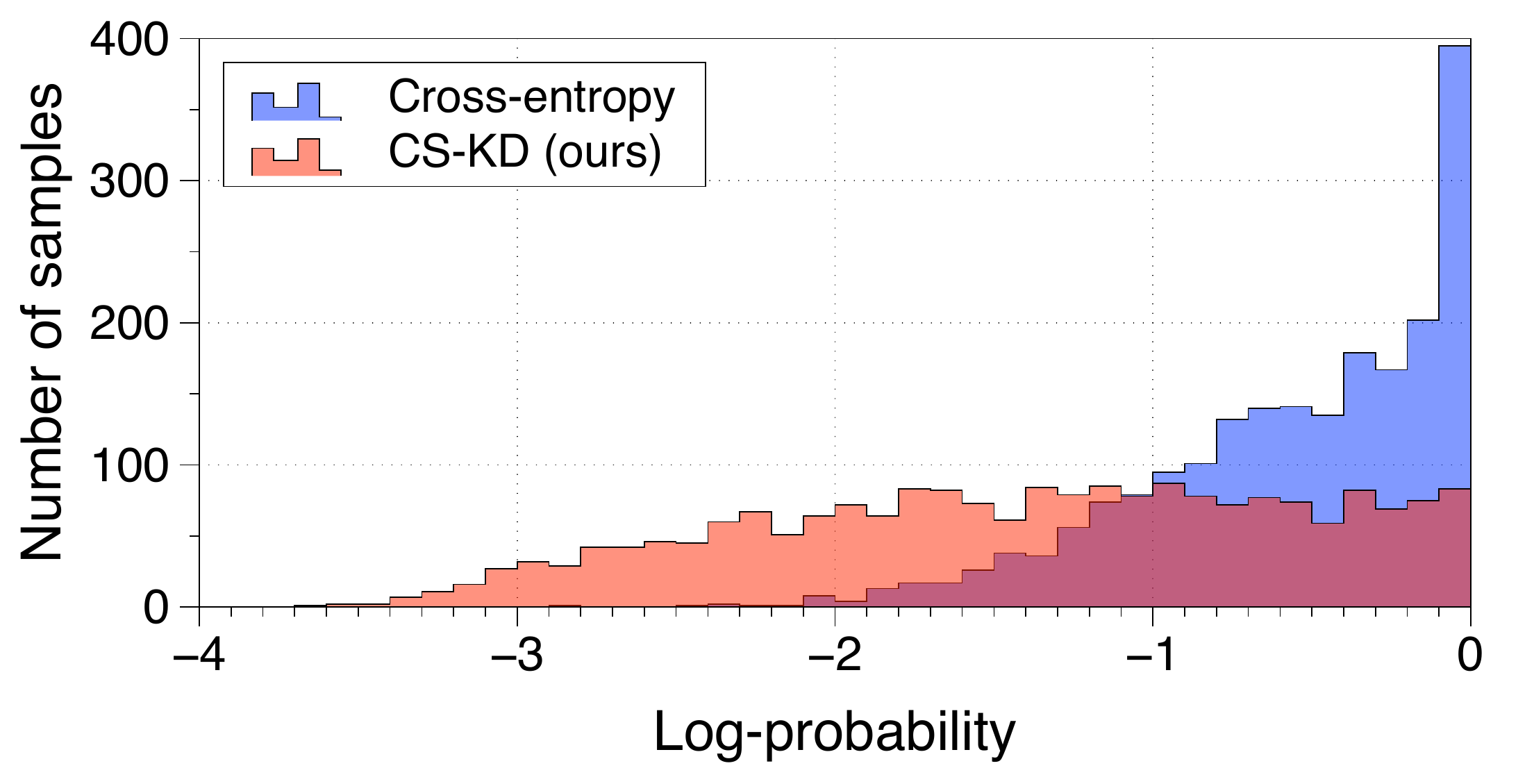}} 
\,
\subfigure[\textcolor{black}{Log-probabilities of ground-truth labels on misclassified samples}]
{
\includegraphics[width=0.46\textwidth]{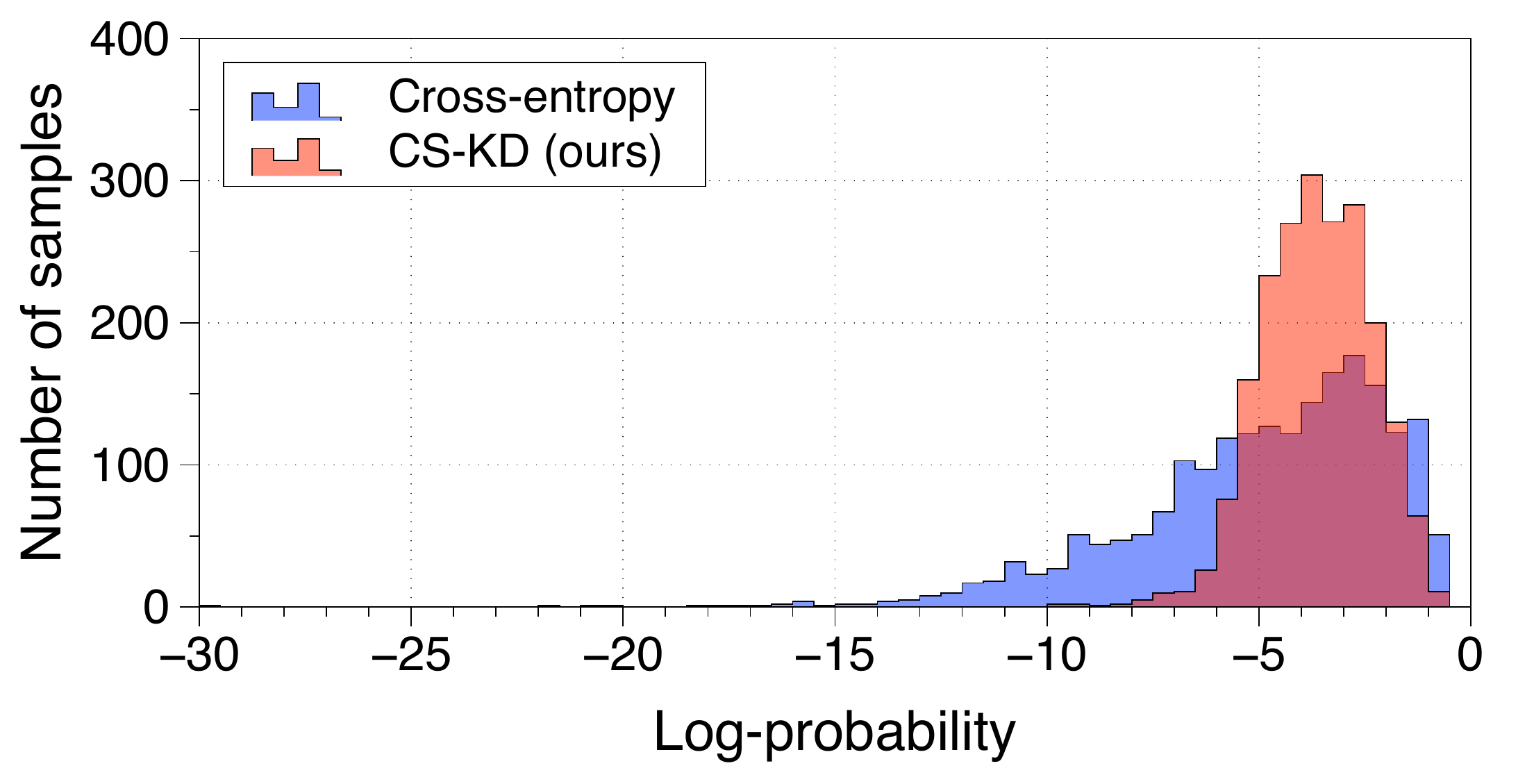}}
\caption{\textcolor{black}{Histogram of log-probabilities of (a) the predicted label, \textit{i.e.}, top-1 softmax score, and (b) the ground-truth label on misclassified samples by networks trained by {the} cross-entropy {(baseline)} and CS-KD.
The networks are trained on ResNet-18 for Stanford Dogs.}}\label{hist:sta}

% \vspace{-0.05in}
\subfigure[\textcolor{black}{Log-probabilities of predicted labels on misclassified samples}]
{
\includegraphics[width=0.46\textwidth]{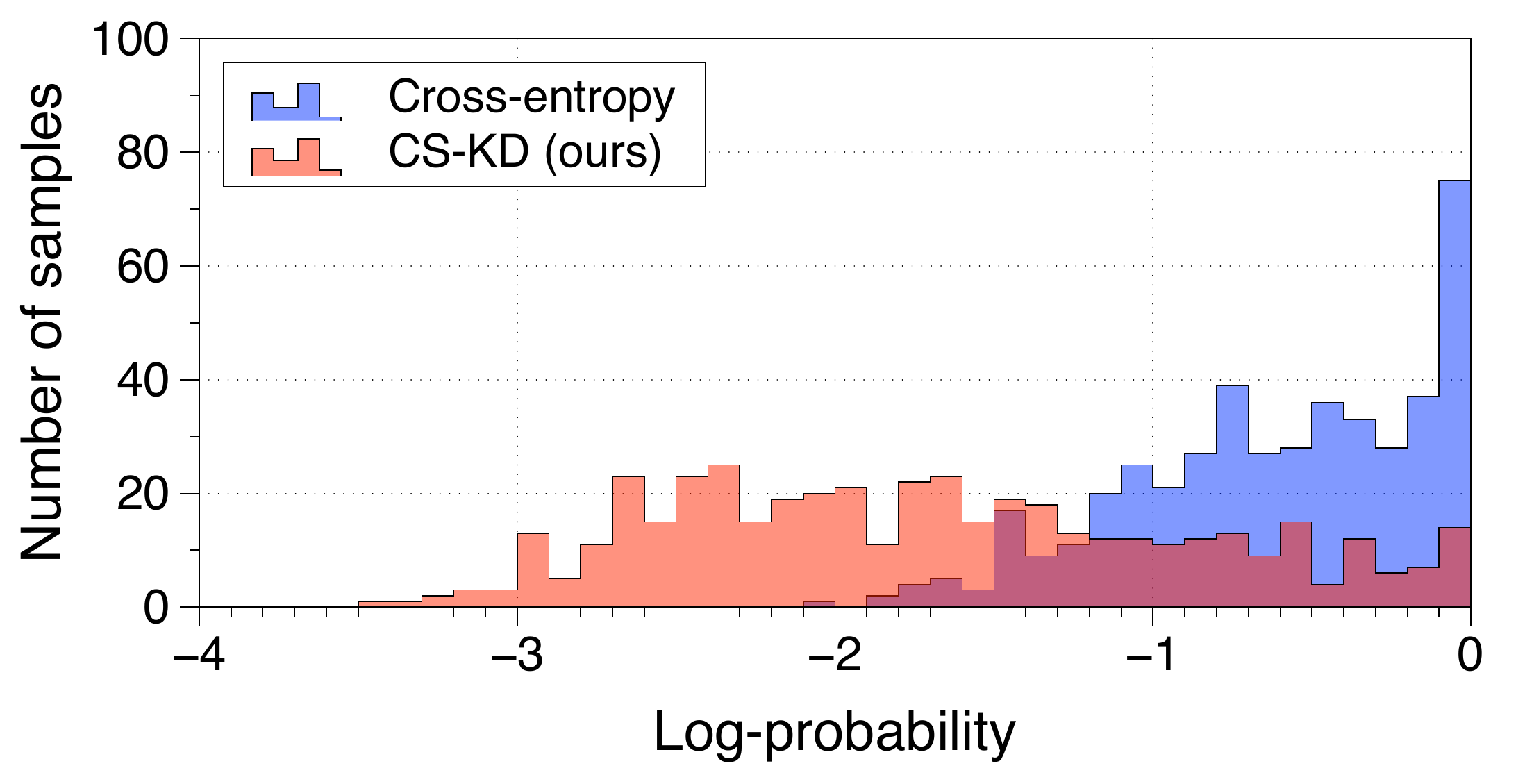}} 
\,
\subfigure[\textcolor{black}{Log-probabilities of ground-truth labels on misclassified samples}]
{
\includegraphics[width=0.46\textwidth]{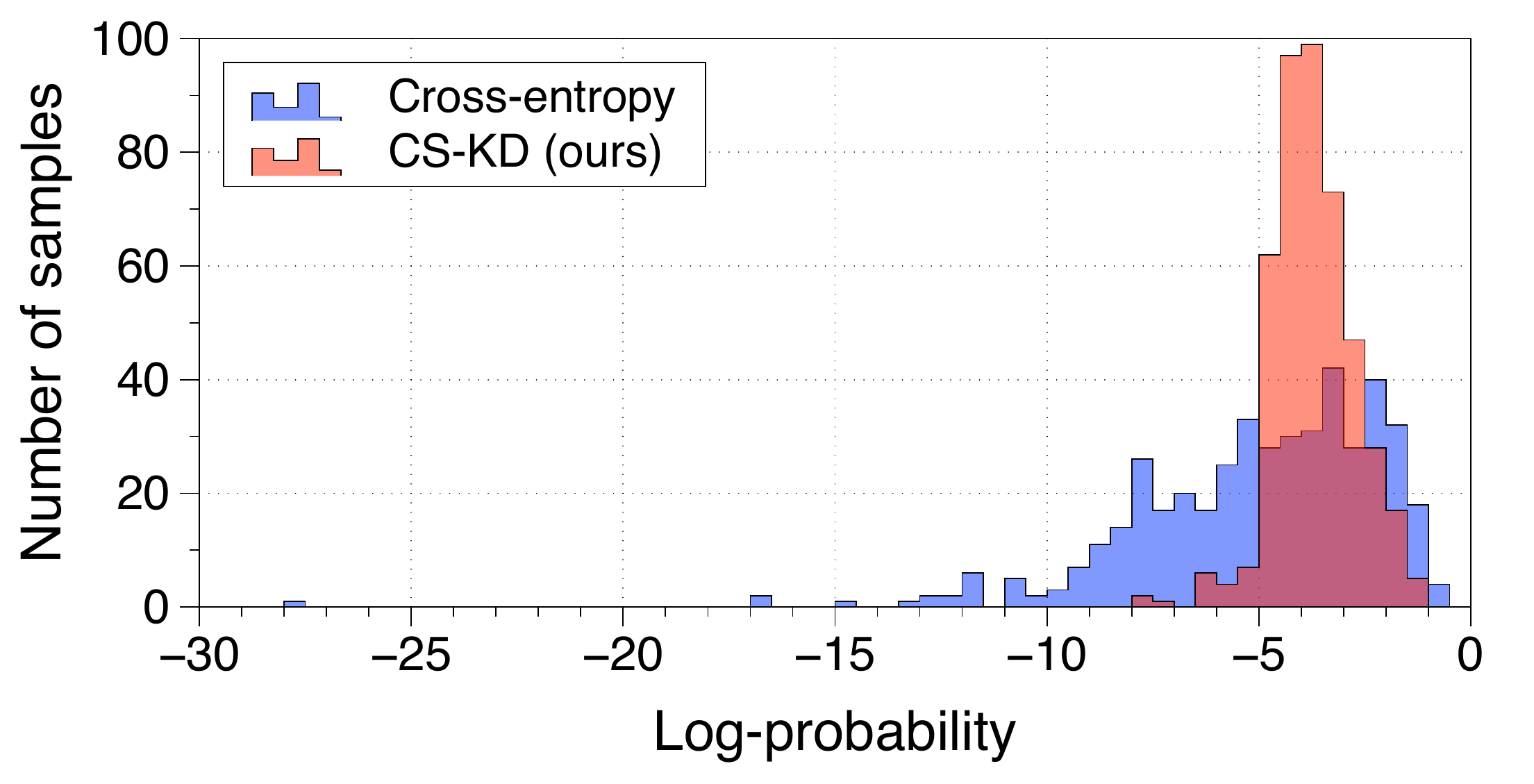}}
\caption{\textcolor{black}{Histogram of log-probabilities of (a) the predicted label, \textit{i.e.}, top-1 softmax score, and (b) the ground-truth label on misclassified samples by networks trained by {the} cross-entropy {(baseline)} and CS-KD.
The networks are trained on ResNet-18 for MIT67.}}\label{hist:mit}
\end{figure*}

Moreover, we additionally compare our method with the cross-entropy method by plotting log-probabilities of the softmax scores on commonly misclassified samples for TinyImageNet, CUB-200-2011, Stanford Dogs, and MIT67 datasets.
The corresponding results are reported in Figures~\ref{hist:tiny}, \ref{hist:cub}, \ref{hist:sta}, and \ref{hist:mit}. 
Log-probabilities of the softmax scores on the predicted class show how overconfident the predictions are, and our method produces less confident predictions on the misclassified samples compared to the cross-entropy method for overall datasets.
On the other hand, log-probabilities of the softmax scores on the ground-truth class show relations between the predictions and the ground-truth class, and our method increases the ground-truth scores for overall datasets.
% are commonly misclassified by both the cross-entropy and our method for TinyImageNet, CUB-200-2011, Stanford Dogs and MIT67 datasets.
% We use PreAct ResNet-18 for TinyImagenet, and ResNet-18 for all of fine-grained datasets.
% We emphasize that our method not only produces less confident predictions on misclassified samples compared to the cross-entropy method but also
% increases ground-truth scores for overall datasets.
These results imply that our method induces meaningful predictions that are more related to the ground-truth class than the cross-entropy method.

% \begin{figure*}[h]
% \centering
% % \vspace{-0.30in}
% \subfigure[{Log-probabilities of predicted labels on misclassified samples}]
% {
% \includegraphics[width=0.46\textwidth]{figures/Tiny_r18_log_py_max.pdf}} 
% \,
% \subfigure[{Log-probabilities of ground-truth labels on misclassified samples}]
% {
% \includegraphics[width=0.46\textwidth]{figures/Tiny_r18_log_py_true.pdf}}
% \caption{\textcolor{black}{Histogram of log-probabilities of (a) the predicted label, \textit{i.e.}, top-1 softmax score, and (b) the ground-truth label on misclassified samples by networks trained by {the} cross-entropy {(baseline)} and CS-KD.
% The networks are trained on PreAct ResNet-18 for TinyImageNet.}}\label{hist:tiny}
% \end{figure*}

\end{document}